\documentclass{article}

\usepackage[preprint]{neurips_2026}
\usepackage[utf8]{inputenc}
\usepackage[T1]{fontenc}
\usepackage{hyperref}
\usepackage{url}
\usepackage{booktabs}
\usepackage{amsfonts}
\usepackage{amsmath}
\usepackage{amssymb}
\usepackage{microtype}
\usepackage{graphicx}
\usepackage{xcolor}
\usepackage{subcaption}
\usepackage{multirow}
\usepackage{enumitem}
\hypersetup{hidelinks}

\newcommand{\medtok}{\textsc{medtokenizers}}
\newcommand{\medlat}{\textsc{medlatents}}

\title{Tokenizer--Generator Coupling in Medical Image Generation}

\author{%
  Liam Chalcroft \\
  University College London \\
  London, United Kingdom \\
  \texttt{liam.chalcroft.20@ucl.ac.uk} \\
}

\begin{document}

\maketitle

\begin{abstract}
Latent medical image generators usually treat the tokenizer as fixed preprocessing. We test whether this separation holds in a controlled ChestMNIST study at $64\times64$ that crosses discrete tokenizers, generator families, and sampler settings under a shared latent grid, with continuous-latent reference cells. The rankings depend jointly on the tokenizer, generator, and sampler. The best quantizer changes with the generator, and validation-based sampler selection changes the apparent generator ranking. We interpret this through a rate-distortion-modelability framing in which modelability is conditional on the generator, sampler, and inference budget. The interaction persists when the vocabulary-1024 block is retrained at three seeds, with 4 of 9 pairwise quantizer comparisons exceeding three seed standard deviations, including a reversal between LFQ and FSQ from MaskGIT to D3PM; the rest of the grid was trained at a single seed and is correspondingly less certain. Reconstruction PSNR alone is not a reliable selection criterion. On LFQ-1024, retuning the D3PM and SEDD samplers on a held-out validation split reduces FID-192 from 0.44/0.41 at the default budget to 0.09/0.10 at lower NFE, replicated across seeds, although the continuous references were not given an equivalent sampler sweep. FID-192, our internal ranking metric, ranks consistently with standard FID-2048 (Spearman 0.89) and with a label-free classifier two-sample test (0.86). All experiments are unconditional, use low-resolution $64\times64$ medical-style images and are evaluated with non-clinical FID-based metrics, and our claims are limited to that setting. Implementations are released at \url{https://github.com/liamchalcroft/medtokenizers} and \url{https://github.com/liamchalcroft/medlatents}.
\end{abstract}

\section{Introduction}
\label{sec:intro}

Generative modelling of medical images is a methodological tool for studying data augmentation~\citep{prakash2024augmentation}, synthetic-data workflows that can improve downstream classifier fairness~\citep{ktena2024fairness}, and rare-condition modelling. Diffusion-based approaches now dominate, with latent diffusion models achieving strong image quality for chest X-rays~\citep{chambon2024roentgen}, brain MRI~\citep{khader2023ddpm3d}, and histopathology~\citep{yellapragada2024pathldm}. These systems usually inherit a narrow design from natural-image synthesis, in which a VQ-style or KL-regularised autoencoder tokenizer is followed by continuous Gaussian diffusion.

Computer vision now offers alternatives along two axes. On the tokenizer side, vector quantization (VQ)~\citep{vqvae}, finite scalar quantization (FSQ)~\citep{mentzer2024fsq}, lookup-free quantization (LFQ)~\citep{yu2024magvitv2}, and continuous VAE/AE variants can all produce usable reconstructions while imposing different latent distributions. On the generator side, autoregressive transformers, masked generative models~\citep{chang2022maskgit}, discrete diffusion~\citep{austin2021d3pm}, score-entropy discrete diffusion~\citep{lou2024sedd}, discrete flow matching~\citep{gat2024dfm,campbell2024dfm}, Bayesian Flow Networks~\citep{graves2023bfn}, latent diffusion~\citep{rombach2022ldm}, and rectified flow~\citep{liu2023flow} encode different assumptions about how a latent distribution should be sampled. However, to our knowledge no prior medical-image study has evaluated the interaction between these two design axes under a common protocol.

Recent evidence from natural images shows a \emph{reconstruction--generation tradeoff}, in which improving tokenizer reconstruction can degrade generation quality~\citep{yao2025vavae,xiong2025gigatok,ramanujan2024crt}. Extending the classical rate-distortion-perception tradeoff in lossy compression~\citep{blau2019rdp}, \citet{dieleman2025rdm} frames this as a three-way \emph{rate-distortion-modelability} tradeoff, where the optimal tokenizer balances compression fidelity against how easily the compressed representation can be modelled by a generator. Whether this principle holds for medical images, where anatomical constraints reduce visual diversity relative to natural images, is unknown.

This work is a controlled factorial study in which the unit of study is the
tokenizer--generator--sampler triple. A controlled ChestMNIST-64 setting is
used to expose design interactions that are difficult to observe in
full-scale pipelines, and the libraries are released so that the same
comparisons can be repeated on higher-resolution, 3D, or
institution-specific datasets.

The implementation is split into two repositories. \medtok{} (\url{https://github.com/liamchalcroft/medtokenizers}) contains the tokenizer training, reconstruction, and tokenization stack. \medlat{} (\url{https://github.com/liamchalcroft/medlatents}) contains the latent-generator training, sampling, and evaluation stack. 

This work makes four contributions:
\begin{enumerate}
    \item A controlled empirical study of latent generative design for medical images, comprising 70 ChestMNIST-64 generation cells, of which 54 are discrete tokenizer-generator cells from a matched VQ/LFQ/FSQ vocabulary sweep and 16 are continuous-latent reference cells from VAE/AE channel and KL sweeps.
    \item An evaluation of underexplored latent generator families (autoregressive transformers, MaskGIT, DFM, D3PM, SEDD, BFN, LDM, and RF) under a shared medical-image protocol.
    \item Empirical findings on tokenizer-generator-sampler interaction. Within a seed-verified vocabulary-1024 block the best quantizer changes with the generator, so the two are not separable; reconstruction quality is a poor proxy for generation quality; and the best D3PM/SEDD sampling budgets depend on the tokenizer. We also propose a selection protocol in which the tokenizer is evaluated jointly with the generator and sampler, with reconstruction quality not used as the selection criterion.
    \item Extensible libraries, \medtok{} and \medlat{}, which support 2D/3D medical-image tokenization and latent generation and include the FID-192 bootstrap, domain-FID, throughput, factor-decomposition, and nearest-neighbour memorization utilities used in this study.
\end{enumerate}

\begin{table}[h]
\centering
\caption{Controlled study design.}
\label{tab:study_design}
\small
\begin{tabular}{@{}p{0.28\linewidth}p{0.64\linewidth}@{}}
\toprule
\textbf{Component} & \textbf{Experimental protocol} \\
\midrule
Primary task & Unconditional generation of medical images at $64 \times 64$ \\
Primary dataset & ChestMNIST at $64 \times 64$; 78K training images and 22.4K test images \\
Replication datasets & PneumoniaMNIST and OrganAMNIST at $64 \times 64$ \\
Discrete panel & VQ/LFQ/FSQ $\times$ vocabularies 1{,}024/2{,}048/4{,}096 $\times$ six discrete generators = 54 cells \\
Continuous references & Eight distinct VAE/AE settings spanning channel and KL sweeps (the $c\!=\!4$, $\lambda_\text{KL}\!=\!10^{-6}$ corner is shared by the two sweeps, so it is one setting evaluated once), each with LDM/RF = 16 cells \\
Seeds and samples & One training seed per cell for the full grid; the vocabulary-1024 interaction block re-run at three seeds (42/43/44); 10K samples per default cell; sampler hyperparameters screened on 2K validation samples, with selected configurations evaluated once on 10K test samples \\
Primary metric & FID-192 as an internal ranking metric, with bootstrap uncertainty \\
Secondary checks & Domain-FID sanity check, reconstruction metrics, throughput, cross-dataset replication, and nearest-neighbour memorization probe \\
\bottomrule
\end{tabular}
\end{table}

\section{Related Work}
\label{sec:related}

\paragraph{Medical image tokenization.}
Most medical-image generation systems still use a narrow set of tokenizers: VQ-style autoencoders~\citep{tschannen2018autoencoder}, KL-regularised continuous autoencoders, or task-specific continuous latent encoders. MONAI Generative Models~\citep{pinaya2023monai} provides VQ-VAE and AutoencoderKL, and MedVAE~\citep{varma2025medvae} focuses on continuous autoencoder latents and does not include a matched discrete-quantizer comparison; MedITok~\citep{ma2025meditok} unifies a medical-image tokenizer for synthesis and interpretation but likewise does not treat the quantizer family as a controlled variable. Newer quantization strategies (FSQ~\citep{mentzer2024fsq} and LFQ~\citep{yu2024magvitv2}, alongside variational discrete tokenization~\citep{yang2025vaevq} and high-utilisation codebook scaling~\citep{zhu2024vqganlc}) have rarely been evaluated systematically for medical image generation. The choice of tokenizer is relevant to medical imaging because it determines which anatomical information is preserved and which latent distribution the generator must learn.

\paragraph{Discrete generative models.}
Autoregressive transformers~\citep{esser2021vqgan} and masked generative models~\citep{chang2022maskgit,yu2024magvitv2,weber2024maskbit} are natural choices once an image has been turned into a sequence of categorical tokens, and recent natural-image work scales this approach~\citep{sun2024llamagen}, reorders it as next-scale prediction~\citep{tian2024var}, or removes vector quantization altogether~\citep{li2024mar}. Discrete diffusion (D3PM~\citep{austin2021d3pm} and SEDD~\citep{lou2024sedd}), simplified and generalised masked diffusion~\citep{shi2024md4}, discrete flow matching~\citep{gat2024dfm,campbell2024dfm}, and BFN~\citep{graves2023bfn} offer further alternatives with different sampling dynamics. In medical imaging, discrete generators beyond VQ-GAN-style compression remain underexplored.

\paragraph{Tokenizer--generator interaction.}
\citet{yu2024magvitv2} demonstrated that upgrading only the tokenizer (VQ to LFQ) can improve generation with the same generator. \citet{yao2025vavae} showed that increasing latent capacity improves reconstruction but can worsen generation. Recent tokenizer benchmarks such as VTBench~\citep{lin2025vtbench} and TokBench~\citep{tokbench2025} likewise argue that reconstruction metrics do not settle tokenizer quality for generation. Factorized Quantization (FQGAN)~\citep{fqgan2024} sweeps multiple quantization variants but holds the generator fixed; codebook scaling with high utilisation~\citep{zhu2024vqganlc} and tokenizer post-training~\citep{robustok2025} further indicate that the tokenizer, alongside the generator, is a primary determinant of generation quality. This work studies the interaction on medical data and crosses six discrete generator families with the matched-vocabulary 3$\times$3 quantization grid, which prior natural-image studies do not.

\paragraph{MedMNIST generation.}
Despite MedMNIST's status as a standard classification benchmark~\citep{yang2023medmnist}, we are aware of no standardized, multi-architecture generative protocol for MedMNIST. Published generation results are sparse and limited to unconditional Generative Adversarial Networks (GANs) and basic Denoising Diffusion Probabilistic Models (DDPMs) at $28 \times 28$, with ChestMNIST-$64 \times 64$ generation metrics largely absent.

\paragraph{Evaluating medical image generation.}
Generation quality is most often summarised by FID~\citep{heusel2017fid} and Inception Score~\citep{salimans2016is}, both of which inherit an ImageNet-trained feature space that transfers imperfectly to medical images. \citet{woodland2024miccai} examine these feature-extractor assumptions for medical generative models, and the Fr\'echet Radiomic Distance~\citep{konz2024frd} substitutes radiomic features as a domain-aware alternative. More recently, {CheXGenBench}~\citep{dutt2025chexgenbench} assembles a unified chest-radiograph benchmark spanning fidelity, privacy, and downstream utility, although it ranks end-to-end text-to-image systems and does not isolate the tokenizer, generator, and sampler choices studied here. We adopt a low-resolution FID variant as an internal ranking metric and cross-check it against a domain-trained feature space (\S\ref{sec:eval}, App.~\ref{app:fid_validation}), so that no single feature extractor is treated as ground truth.

\section{Controlled Study Design}
\label{sec:methods}

Table~\ref{tab:study_design} summarises the design. The study
is a factorial sweep over three design axes (the quantizer family and vocabulary size of the tokenizer, and the generator family that models the resulting latents) evaluated under one protocol, with the sampler varied in a separate sweep,
with continuous VAE/AE tokenizers and their LDM/RF generators as reference
cells. Every tokenizer shares the same encoder--decoder backbone, and every
generator is trained and evaluated through the same pipeline; generator
capacity, learning rate, and training length are architecture-specific and
reported in \S\ref{sec:generators}. Because these details differ, the experimental unit is the full triple.

\subsection{Image Tokenizers}
\label{sec:tokenizers}

All tokenizers share an identical convolutional encoder-decoder backbone with $8\times$ spatial compression~\citep{esser2021vqgan,rombach2022ldm}, mapping $64 \times 64$ greyscale chest X-rays to $8 \times 8$ latent grids. The backbone uses channel dimensions $(64, 128, 256)$, two residual blocks per resolution, single-headed self-attention at $16 \times 16$, and is trained with L1 reconstruction loss (weight 4.0), VGG-16 perceptual feature matching loss~\citep{johnson2016perceptual} (weight 0.5), a multi-scale PatchGAN discriminator~\citep{esser2021vqgan} (weight 0.05, hinge loss, starting at epoch 10), and a LeCam regulariser (weight 0.001). Only the quantization layer differs across tokenizers.

Because every
generator in this paper consumes the encoder's output and decodes
through the same decoder, the tokenizer choice largely determines
what the generator is asked to model. We evaluate five tokenizer families that vary
along two axes, discrete vs continuous latents and (for
discrete) the choice of quantization method. Each maps a
$64 \times 64$ image to an $8 \times 8$ spatial grid (64 latent
positions) and differs only in what each position holds.

\paragraph{Vector Quantization (VQ)~\citep{vqvae}.}
VQ is the classical VQ-VAE approach, in which each spatial position outputs a
continuous vector that is replaced by the nearest entry in a
learned codebook of $K$ vectors, and the discrete index of that entry
becomes the token. The codebook entries themselves are updated by an
exponential-moving-average (EMA) rule from the assigned encoder outputs
\citep{vqvae}, so the codebook converges to $K$ representative patch embeddings of the training data. Because the nearest-neighbour assignment is not
differentiable, gradients are copied past it with a straight-through
estimator, and a commitment loss keeps encoder outputs close to their
assigned entries~\citep{vqvae}. Matching uses cosine similarity, so the
selected codebook vector is L2-normalised before decoding, both in training
and when generated tokens are decoded. The learned codebook makes the index
geometry data-adapted but irregular, since nearby indices need not decode to
perceptually similar patches, so the generator must model a less
structured categorical target. VQ remains a common discrete baseline in
medical generative modelling, and its main known failure mode is
\emph{codebook collapse} (a few entries dominate while others are
never used), which we control with codebook normalisation and an
appropriate vocabulary size.

\paragraph{Lookup-Free Quantization (LFQ)~\citep{yu2024magvitv2}.}
The encoder outputs a $d$-dimensional vector at each position, and
the sign of each dimension produces a $d$-bit binary code, with
$2^d$ possible tokens. There is no codebook, as the token vocabulary is
the entire $\{0,1\}^d$ hypercube and the codebook entry for each
token is implicit. To shape this code, an entropy
regulariser~\citep{yu2024magvitv2} lowers each position's assignment
entropy, so that a position commits to a confident bit pattern, whilst
raising the batch-averaged entropy, so that codes are used uniformly
across the dataset. The result is a token distribution close to uniform
over a factorised binary code, which forms a high-entropy categorical
target (Table~\ref{tab:tokenizer_props}). At vocabulary 1{,}024 we use $d\!=\!10$ bits.

\paragraph{Finite Scalar Quantization (FSQ)~\citep{mentzer2024fsq}.}
FSQ is also codebook-free, with the encoder outputting a $d$-dimensional
vector of which each dimension is independently rounded to one of $L_i$ discrete
levels; the joint code is the cartesian product, giving $\prod_i L_i$
tokens. At vocabulary 1{,}024 we use $d\!=\!5$ dimensions of
$L\!=\!4$ levels each ($4^5 = 1024$). Each scalar is first squashed
through a bounded nonlinearity into the quantization range and then
rounded with a straight-through estimator~\citep{mentzer2024fsq}, so FSQ
tiles latent space with a fixed multi-level grid in place of learned code
locations. Because each scalar is rounded
independently, FSQ has no commitment loss and no codebook collapse risk
by construction; the tradeoff is that the per-position distribution
inherits whatever shape the encoder produces, with no entropy
regulariser to flatten it.

\paragraph{Variational Autoencoder (VAE)~\citep{kingma2014vae,rombach2022ldm}.}
The VAE is a common continuous tokenizer in latent diffusion pipelines,
in which the encoder outputs a Gaussian distribution per spatial position
(a mean and a log-variance vector with $c$ channels) and the decoder
operates on a sample. A KL-divergence regulariser pulls each
per-position distribution toward $\mathcal{N}(0,I)$ so the resulting
latent space is approximately unit-Gaussian, which is convenient for
downstream diffusion. Gaussian diffusion and rectified-flow samplers
transport a standard-normal prior to the data, so a latent space already
near $\mathcal{N}(0,I)$ matches the sampler's source distribution and
needs little rescaling, whereas latents whose scale drifts from unity
become harder to sample (App.~\ref{app:tokenizer}). We evaluate two sweeps along this family, the first over
bottleneck capacity ($c \in \{1, 2, 4, 8, 16\}$ at fixed
$\lambda_\text{KL}\!=\!10^{-6}$), which provides continuous storage-proxy
references that are not rate-matched to the discrete rows, and the second
over KL strength
($\lambda_\text{KL} \in \{10^{-5}, 10^{-6}, 10^{-7}\}$ at fixed
$c\!=\!4$), which controls how strongly the latent space is forced
toward $\mathcal{N}(0,I)$.

\paragraph{Autoencoder (AE).}
The AE is the $\lambda_\text{KL}\!=\!0$ endpoint of the continuous VAE/KL sweep.
It uses the same encoder-decoder family as the VAE rows but removes the
force that pulls latents toward a unit Gaussian. AE therefore serves as a
reference point showing which latent scale and geometry best reconstruct
the data when the tokenizer is unconstrained. Without a prior the latent
scale inflates well above unity (App.~\ref{app:tokenizer}), so AE forms the
low-distortion extreme of the continuous family, and its modelability
depends on whether the generator's sampler can absorb that scale. RF handles the resulting latent distribution
well here, while diffusion-style samplers with fixed Gaussian
noise schedules can be more sensitive to scale.

\medskip
Each discrete method is trained at vocabularies of 1{,}024, 2{,}048,
and 4{,}096, forming the controlled $3 \times 3$ comparison that
disentangles quantization method from vocabulary size. Prior
comparisons (including MAGVIT-v2's VQ-to-LFQ upgrade)
confound the two by comparing methods at different vocabulary sizes. We
exclude residual quantization (for example residual FSQ): it stacks
several codes per spatial position to lower distortion, but the resulting
multi-level dependency structure changes what the generator must model
and would confound the single-token-per-position comparison.

\begin{table}[h]
\centering
\caption{Tokenizer reconstruction quality on ChestMNIST test set ($n = 22{,}433$). The $3 \times 3$ discrete grid matches vocabulary sizes across methods. Discrete vocabulary sizes correspond to 640--768 raw token bits at $8 \times 8$ tokens before entropy coding. PSNR = Peak Signal-to-Noise Ratio (dB); SSIM = Structural Similarity Index; LPIPS = Learned Perceptual Image Patch Similarity~\citep{zhang2018lpips}. The continuous block reports the VAE capacity sweep and, below, the KL-strength sweep at fixed $c\!=\!4$ (AE is the $\lambda_\text{KL}\!=\!0$ endpoint).}
\label{tab:tokenizers}
\small
\begin{tabular}{@{}llrrrrr@{}}
\toprule
\textbf{Tokenizer} & \textbf{Type} & \textbf{Vocab} & \textbf{Bits/img} & \textbf{PSNR} & \textbf{SSIM} & \textbf{LPIPS}$\downarrow$ \\
\midrule
VQ-1024   & Discrete & 1{,}024   & 640  & 28.0 & 0.891 & 0.026 \\
VQ-2048   & Discrete & 2{,}048   & 704  & 29.3 & 0.924 & 0.020 \\
VQ-4096   & Discrete & 4{,}096   & 768  & 29.4 & 0.925 & 0.018 \\
\midrule
LFQ-1024  & Discrete & 1{,}024   & 640  & 28.0 & 0.893 & 0.024 \\
LFQ-2048  & Discrete & 2{,}048   & 704  & 28.1 & 0.894 & 0.025 \\
LFQ-4096  & Discrete & 4{,}096   & 768  & 28.6 & 0.904 & 0.021 \\
\midrule
FSQ-1024  & Discrete & 1{,}024   & 640  & 29.4 & 0.912 & 0.020 \\
FSQ-2048  & Discrete & 2{,}048   & 704  & 29.6 & 0.915 & 0.018 \\
FSQ-4096  & Discrete & 4{,}096   & 768  & 29.7 & 0.916 & 0.019 \\
\midrule
VAE-c1    & Continuous & $c\!=\!1$  & 1{,}024\textsuperscript{$\dagger$} & 24.6 & 0.855 & 0.130 \\
VAE-c2    & Continuous & $c\!=\!2$  & 2{,}048\textsuperscript{$\dagger$} & 29.3 & 0.896 & 0.027 \\
VAE-c4    & Continuous & $c\!=\!4$  & 4{,}096\textsuperscript{$\dagger$} & 32.4 & 0.940 & 0.013 \\
VAE-c8    & Continuous & $c\!=\!8$  & 8{,}192\textsuperscript{$\dagger$} & \textbf{34.4} & \textbf{0.963} & \textbf{0.011} \\
VAE-c16   & Continuous & $c\!=\!16$ & 16{,}384\textsuperscript{$\dagger$} & 32.3 & 0.949 & 0.014 \\
\midrule
VAE-1e-5  & Continuous & $\lambda_\text{KL}\!=\!10^{-5}$ & 4{,}096\textsuperscript{$\dagger$} & 28.2 & 0.881 & 0.028 \\
VAE-1e-6  & Continuous & $\lambda_\text{KL}\!=\!10^{-6}$ & 4{,}096\textsuperscript{$\dagger$} & 32.4 & 0.940 & 0.013 \\
VAE-1e-7  & Continuous & $\lambda_\text{KL}\!=\!10^{-7}$ & 4{,}096\textsuperscript{$\dagger$} & 32.0 & 0.937 & 0.014 \\
AE        & Continuous & $\lambda_\text{KL}\!=\!0$       & 4{,}096\textsuperscript{$\dagger$} & 32.3 & 0.940 & 0.012 \\
\bottomrule
\end{tabular}

\vspace{2pt}
{\footnotesize $\dagger$ Storage proxy for continuous tokenizers computed as $c \times 8 \times 8 \times 16$ (float16); this is not an entropy-coded bitrate and is not directly comparable to discrete token entropy.}
\end{table}

\begin{table}[h]
\centering
\caption{Tokenizer properties at vocabulary 1{,}024. All tokenizers share an identical encoder-decoder backbone of 8.5M parameters ($\sim$34\,MB in float32); the quantization head is parameter-light. Code utilisation and marginal token entropy are computed on the test set ($n\!=\!22{,}433$). Entropy is normalised by $\log_2 |V|$. Entropies at vocabularies 2{,}048 and 4{,}096 are given in the text.}
\label{tab:tokenizer_props}
\small
\begin{tabular}{@{}llrrrl@{}}
\toprule
\textbf{Tokenizer} & \textbf{Type} & \textbf{Parameters} & \textbf{Utilisation} & \textbf{Normalised Entropy} & \textbf{Key Property} \\
\midrule
VQ    & Discrete & 8.5M & 100\% & 0.997 & Learned codebook (EMA) \\
LFQ   & Discrete & 8.5M & 100\% & \textbf{0.999} & Binary hypercube \\
FSQ   & Discrete & 8.5M & 100\% & 0.959 & Fixed scalar grid \\
\midrule
VAE   & Continuous & 8.5M & --- & --- & KL-regularised latents \\
\bottomrule
\end{tabular}
\end{table}

LFQ achieves near-perfect marginal token entropy (0.999 normalised; Table~\ref{tab:tokenizer_props}) at every vocabulary tested, meaning its marginal token frequencies are near-uniform, and VQ is close behind (0.997--0.998 across all three vocabularies). In contrast, FSQ is non-monotonic, with a normalised entropy of 0.959 at vocabulary 1{,}024 that climbs to 0.972 at 2{,}048 and then drops to 0.947 at 4{,}096, so FSQ's fixed scalar grid begins to underutilise marginal token entropy at the largest vocabulary, matching FSQ-4096's relative weakness on iterative-diffusion generators in Table~\ref{tab:main_results}. All methods achieve 100\% code utilisation at the vocabulary sizes tested.

We treat vocabulary size as a controlled rate proxy for discrete
tokenizers: at $8 \times 8$ latent positions, vocabularies of 1{,}024,
2{,}048, and 4{,}096 correspond to 640, 704, and 768 raw token bits per
image before entropy coding. These values should not be interpreted as
clinical information content; they define a matched-vocabulary sweep for
comparing discrete quantizer families under a common architecture.
The continuous channel sweep shows that increasing latent storage does not monotonically
improve reconstruction: VAE-c8 achieves the best reconstruction
(PSNR 34.4; Table~\ref{tab:tokenizers}), while VAE-c16 drops to 32.3. The same non-monotonicity
appears on the generation side, where VAE-c8 is the channel sweep's best
reconstructor yet its worst generator (LDM FID 1.61, RF FID 0.32;
Table~\ref{tab:main_results}), while the reconstruction-poor c1 row
ranks in the middle on generation. Tokenizers should therefore be evaluated by downstream generator behaviour as well as by reconstruction.

\subsection{Generative Models}
\label{sec:generators}

All generators share a Diffusion Transformer (DiT) backbone~\citep{peebles2023dit} with 12 layers, hidden dimension 512, and 8 attention heads. AR and MaskGIT omit the timestep modulation pathway (38.9M parameters at vocabulary 1{,}024); the diffusion-style discrete generators (DFM, D3PM, SEDD, BFN) carry the full adaLN-Zero timestep conditioning of \citet{peebles2023dit} (57--58M); the continuous generators (LDM, RF) use a matched ContinuousDiT (57.1M). All generators use Rotary Position Embeddings (RoPE)~\citep{su2021rope} and SiLU activations.

Each generator models $p(\text{token grid})$ so that new grids can be
sampled and decoded through the (frozen) tokenizer to
produce images. The eight generators in this paper differ in how they
factorise that joint distribution and in how many forward passes they
require at inference time. We group them into four families
($N$ denotes the inference step count for our experiments).

\paragraph{Autoregressive and masked token models (discrete).}

\begin{itemize}[leftmargin=*,itemsep=4pt]
    \item \textbf{AR Transformer} ($N\!=\!64$): a decoder-only
    transformer trained with next-token prediction. The 64 tokens of the $8\!\times\!8$ grid are flattened
    in raster-scan order and the network predicts each token
    conditioned on all preceding tokens. At inference, generation is
    sequential, one token per forward pass starting from a beginning-of-sequence symbol, with key-value caching
    to amortise the prefix computation. It
    factorises the joint exactly as $p(x)=\prod_i p(x_i \mid x_{<i})$ and
    is trained by minimising the per-token negative log-likelihood, which
    makes it the only exact-likelihood, strictly sequential generator
    among the eight.

    \item \textbf{MaskGIT} ($N\!=\!12$)~\citep{chang2022maskgit}: a
    bidirectional alternative trained with a BERT-style
    masked-language-model objective, in which a random subset of tokens is
    replaced by a special $[\text{MASK}]$ symbol and the network
    predicts the originals from the visible context. At inference, all
    64 positions start masked; over 12 iterative steps the network
    proposes tokens at every still-masked position and a
    cosine-schedule policy commits the highest-confidence subset
    (confidence annealed with Gumbel noise~\citep{maddison2017concrete}), repeating
    until all positions are committed. It is faster than AR ($N\!=\!12$ vs
    64) and uses bidirectional context, at the cost of slightly higher
    per-step memory. Unlike AR it does not define a single consistent
    joint over tokens: each step draws from per-position marginals
    conditioned on the visible set, so the
    AR-vs-MaskGIT contrast compares strict sequential ordering with
    parallel, confidence-ordered commitment.
\end{itemize}

\paragraph{Iterative discrete diffusion / flow.}

\begin{itemize}[leftmargin=*,itemsep=4pt]
    \item \textbf{D3PM (Discrete Denoising Diffusion Probabilistic
    Model)} ($N\!=\!1{,}000$)~\citep{austin2021d3pm}: the categorical
    analogue of continuous Gaussian diffusion. A forward
    Markov chain gradually corrupts each token by replacing it with a
    $[\text{MASK}]$ symbol with probability that grows over $T$
    timesteps following a cosine schedule. The reverse process
    is a learned transformer that, given the partly-masked sequence
    at timestep $t$, predicts the original tokens; a closed-form
    posterior turns that prediction into the slightly-less-corrupted
    sequence at timestep $t-1$. Training maximises a variational lower
    bound on the token-sequence log-likelihood, and we report the
    standard 1{,}000-step budget as the default, although
    \S\ref{sec:vocab} shows that the best tuned budget is far smaller
    and tokenizer-dependent. Absorbing transitions are preferred over
    uniform because they yield a corruption process analogous to
    masked language modelling. We contribute a matrix-free
    implementation (Appendix~\ref{app:implementation}) that reduces
    transition-matrix storage from $O(TK^2)$ to $O(T)$, enabling D3PM training with
    vocabularies up to $K\!=\!64{,}000$ on a single 48\,GB GPU.

    \item \textbf{SEDD (score-entropy loss, D3PM sampler)}
    ($N\!=\!1{,}000$): shares D3PM's forward corruption process, DiT backbone and reverse sampling procedure, and is trained with the concrete-score (denoising score entropy)
    objective of \citet{lou2024sedd}, which models the ratios
    $p_t(y)/p_t(x)$ between neighbouring token states in place of a
    variational lower bound. Our implementation uses a simplified
    constant-target formulation with a hybrid cross-entropy
    regulariser ($\lambda\!=\!0.001$) for training stability. Canonical
    SEDD samples by tau-leaping over the learned score; we instead use
    D3PM's closed-form reverse step, so
    the SEDD-vs-D3PM comparison isolates the training objective.

    \item \textbf{Discrete Flow Matching (DFM)}
    ($N\!=\!100$)~\citep{gat2024dfm}: a continuous-time framing of
    discrete generation. Instead of a discrete-time Markov chain it
    defines a probability path from a noise distribution over tokens
    (here a $[\text{MASK}]$-only distribution) to the data
    distribution, parameterised by polynomial schedules ($n\!=\!3$),
    and trains a network to predict the probability velocity (the rate of
    the token-jump process) at any point along the path. At inference, an
    Euler discretisation of the resulting continuous-time Markov chain
    simulates token jumps in 100 steps, an order of magnitude fewer
    than D3PM/SEDD. Against D3PM this contrasts a continuous-time discrete
    flow with a discrete-time absorbing Markov chain over a related
    mask-to-data path.
\end{itemize}

\paragraph{Iterative Bayesian update on the simplex.}

\begin{itemize}[leftmargin=*,itemsep=4pt]
    \item \textbf{BFN (Bayesian Flow Network)}
    ($N\!=\!1{,}000$)~\citep{graves2023bfn}: a model conceptually distinct
    from the diffusion lineage, in which the state of the system at each
    step is a categorical distribution over tokens at every spatial
    position (for a vocabulary of size $K$, an $8\!\times\!8\!\times\!K$
    probability tensor). During training a noisy sender distribution is formed
    from the data and the network is trained to predict a receiver
    distribution, with a continuous-time loss comparing the two on the
    simplex~\citep{graves2023bfn}. At generation time the
    simplex state starts from the prior and is iteratively refined by a
    Bayesian update driven by the network's own receiver distribution. The
    overall scale of the accuracy schedule is left to heuristics in the
    original work; we set it from the vocabulary as $\beta\!=\!\sqrt{2 \ln K}$
    so the sender distribution concentrates sufficient mass on the correct
    class by the final timestep. BFN trains efficiently but requires a large step budget
    at inference and, in our experiments, exhibits a
    higher FID floor than the diffusion-style alternatives.
\end{itemize}

\paragraph{Continuous-latent generators (consume VAE / AE latents).}

\begin{itemize}[leftmargin=*,itemsep=4pt]
    \item \textbf{LDM (Latent Diffusion Model)}
    ($N\!=\!1{,}000$)~\citep{rombach2022ldm,nichol2021improved}:
    standard continuous Gaussian diffusion in the VAE latent space.
    The forward process adds incremental Gaussian noise over $T$
    timesteps (cosine schedule), and the reverse process is a learned
    ContinuousDiT predicting the noise residual at each step. This
    setup is closest to prior medical-imaging diffusion work
    (e.g.\ RoentGen~\citep{chambon2024roentgen} for chest X-rays).
    Sampling uses stochastic ancestral (DDPM) steps;
    paired with RF on identical VAE latents, the LDM-vs-RF comparison is
    a recipe-level contrast between this diffusion setup and RF's
    deterministic ODE integration.

    \item \textbf{RF (Rectified Flow)}
    ($N\!=\!100$)~\citep{liu2023flow}: a continuous-time flow-matching
    alternative for the same VAE latent space, which trains a velocity
    field to predict the straight-line direction from noise to data
    at any interpolation, then samples by integrating the resulting
    ODE. Because these target paths are near-straight, coarse ODE
    discretisation adds little error, so RF uses fewer sampler
    steps than LDM (100 ODE steps vs 1{,}000 DDPM steps).
    It matches or improves on LDM's FID-192 on every cell
    except $c\!=\!1$, $c\!=\!2$, and $\lambda_\text{KL}\!=\!10^{-7}$ (Table~\ref{tab:main_results}).
\end{itemize}

For continuous generators, latents can optionally be standardised. When
latent normalisation is enabled, the mean and standard deviation are
estimated on the training split, generated latents are un-normalised
before decoding. This applies in particular to AE, whose raw latent scale is
not constrained by a KL prior; in the ChestMNIST runs AE and VAE-c1 use
latent normalisation, while the other VAE channel and KL-sweep
rows use their raw tokenizer latents.

All models are trained for 100 epochs (200 for continuous generators) with AdamW ($\text{lr} = 3 \times 10^{-4}$ for AR/MaskGIT/DFM, $\text{lr} = 10^{-4}$ for D3PM/SEDD/BFN/LDM/RF), cosine learning rate schedule with 5\% warmup, EMA~\citep{polyak1992averaging} ($\alpha = 0.9999$), gradient clipping at 1.0, BFloat16 mixed precision, and batch size 128 on a single NVIDIA A6000 (48\,GB). Table~\ref{tab:generator_settings} collects the per-generator settings that vary across families.

\begin{table}[h]
\centering
\caption{Per-generator training and sampling settings. All generators share the DiT-style backbone (12 layers, width 512, 8 heads); parameter counts differ because the diffusion-style and continuous variants carry adaLN-Zero timestep conditioning. Optimiser settings and hardware are shared across generators (\S\ref{sec:generators}).}
\label{tab:generator_settings}
\small
\begin{tabular}{@{}llrrrrl@{}}
\toprule
\textbf{Generator} & \textbf{Family} & \textbf{Params} & \textbf{LR} & \textbf{Epochs} & \textbf{Steps $N$} & \textbf{Latent norm} \\
\midrule
AR Transformer & Discrete   & 38.9M   & $3\times10^{-4}$ & 100 & 64    & --- \\
MaskGIT        & Discrete   & 38.9M   & $3\times10^{-4}$ & 100 & 12    & --- \\
DFM            & Discrete   & 57--58M & $3\times10^{-4}$ & 100 & 100   & --- \\
D3PM           & Discrete   & 57--58M & $10^{-4}$        & 100 & 1{,}000 & --- \\
SEDD        & Discrete   & 57--58M & $10^{-4}$        & 100 & 1{,}000 & --- \\
BFN            & Discrete   & 57--58M & $10^{-4}$        & 100 & 1{,}000 & --- \\
LDM            & Continuous & 57.1M   & $10^{-4}$        & 200 & 1{,}000 & AE, VAE-c1 \\
RF             & Continuous & 57.1M   & $10^{-4}$        & 200 & 100   & AE, VAE-c1 \\
\bottomrule
\end{tabular}
\end{table}

\subsection{Evaluation Protocol}
\label{sec:eval}

For each trained model, we generate 10{,}000 samples using EMA weights, decode them to pixel space through the corresponding frozen tokenizer decoder, and compute FID~\citep{heusel2017fid} against the 22{,}433 real test images.

\paragraph{FID at low resolution.} Standard FID uses InceptionV3's 2{,}048-dimensional final average-pooling features. We instead compute FID-192, the globally averaged features from the second max-pooling block, via \texttt{FrechetInceptionDistance(feature=192, normalize=True)} in \texttt{torchmetrics}~\citep{torchmetrics}. FID-192 is used for convenience in the low-resolution regime, although standard FID-2048 is also usable at our sample size (its real-vs-real floor at 10K samples is 2.31, std 0.016, with no negative values across 20 resamples; App.~\ref{app:fid_validation}), and FID-192 ranks consistently with it (Spearman $\rho = 0.89$ across 12 cells spanning the full quality range). Because FID-192 reads a lower feature layer, its absolute values are not comparable to standard FID-2048 numbers. We additionally sanity-check the rankings against a domain-relevant feature space (a ResNet-18 multi-label ChestMNIST classifier, mean test AUC 0.761) in App.~\ref{app:fid_validation} and find Spearman rank-correlation $\rho = 0.71$ (exact $p=0.14$) between FID-192 and domain-FID across the six representative ChestMNIST cells used for cross-dataset replication, which supports the coarse FID-192 ranking but is too small to confirm it alone, and does not validate fine-grained orderings or the tuned sampling results.

\paragraph{FID estimator noise floor.} To bound metric-estimator noise we ran a real-vs-real bootstrap on the 22{,}433-image ChestMNIST test set: for each sample size $n \in \{1{,}000, 2{,}500, 5{,}000, 10{,}000\}$, we drew $B=200$ random splits into two disjoint halves of size $n$ each and computed FID-192 between them. Because both halves come from the same distribution, the resulting FID distribution characterises the minimum detectable FID gap at each sample size under this feature extractor. At our 10{,}000-sample setting the FID-192 estimator 95\,\% CI is $[0.001, 0.004]$ (mean 0.002, std 0.001); at 1{,}000 samples it widens to $[0.008, 0.039]$. This bootstrap quantifies estimator uncertainty for fixed generated sample sets. It does not capture training-seed variance or model-selection uncertainty, so we use it only to separate coarse quality regimes. Per-cell estimator CIs from a paired real-versus-generated bootstrap are reported for representative results in App.~\ref{app:fid_ci}. Both bootstrap procedures are included in \medlat{}.

\section{Results}
\label{sec:results}

\subsection{Default Behaviour Shows Generator, Quantizer, and Interaction Effects}
\label{sec:vocab}

Table~\ref{tab:main_results} reports generation FID
across the 54 discrete tokenizer--generator cells and 16 continuous-latent
reference cells.

\begin{table}[t]
\centering
\caption{Generation quality (FID-192 $\downarrow$, 10K samples) across tokenizer--generator cells. The $3 \times 3$ discrete grid (top) enables controlled comparison of quantization methods at matched vocabulary sizes. The continuous block (bottom) sweeps bottleneck capacity ($c\!=\!1\ldots16$ at fixed $\lambda_\text{KL}\!=\!10^{-6}$) and KL regularisation strength ($\lambda_\text{KL}\!=\!10^{-5}\ldots 0$ at fixed $c\!=\!4$, with AE denoting the $\lambda_\text{KL}=0$ endpoint). VAE-$10^{-6}$ and VAE-c4 are the shared $c\!=\!4$, $\lambda_\text{KL}\!=\!10^{-6}$ corner of the two sweeps, so the continuous panel spans eight distinct settings (16 cells). Bold marks the best value in each row.}
\label{tab:main_results}
\small
\begin{tabular}{@{}lr|cccccc|cc@{}}
\toprule
& & \multicolumn{6}{c|}{\textbf{Discrete Generators}} & \multicolumn{2}{c}{\textbf{Continuous}} \\
\textbf{Tokenizer} & \textbf{Vocab} & AR & MaskGIT & DFM & D3PM & SEDD & BFN & LDM & RF \\
\midrule
VQ-1024   & 1K   & \textbf{0.52} & 1.86 & 4.42 & 1.30 & 1.31 & 5.98 & --- & --- \\
VQ-2048   & 2K   & \textbf{0.52} & 3.33 & 2.75 & 2.12 & 2.13 & 3.94 & --- & --- \\
VQ-4096   & 4K   & \textbf{0.31} & 2.03 & 3.39 & 3.69 & 3.54 & 4.22 & --- & --- \\
\midrule
LFQ-1024  & 1K   & \textbf{0.33} & 1.91 & 1.77 & 0.44 & 0.41 & 2.27 & --- & --- \\
LFQ-2048  & 2K   & \textbf{0.45} & 3.40 & 1.33 & 0.74 & 1.02 & 1.46 & --- & --- \\
LFQ-4096  & 4K   & \textbf{0.41} & 1.27 & 2.21 & 1.92 & 2.08 & 2.41 & --- & --- \\
\midrule
FSQ-1024  & 1K   & \textbf{0.39} & 1.15 & 3.02 & 1.21 & 1.05 & 7.60 & --- & --- \\
FSQ-2048  & 2K   & \textbf{0.73} & 1.48 & 1.86 & 2.50 & 2.46 & 4.83 & --- & --- \\
FSQ-4096  & 4K   & \textbf{0.37} & 1.46 & 2.69 & 7.40 & 7.34 & 8.46 & --- & --- \\
\midrule
VAE-c1    & $c\!=\!1$   & --- & --- & --- & --- & --- & --- & \textbf{0.22} & 0.26 \\
VAE-c2    & $c\!=\!2$   & --- & --- & --- & --- & --- & --- & \textbf{0.14} & 0.18 \\
VAE-c4    & $c\!=\!4$   & --- & --- & --- & --- & --- & --- & 0.41 & \textbf{0.13} \\
VAE-c8    & $c\!=\!8$   & --- & --- & --- & --- & --- & --- & 1.61 & \textbf{0.32} \\
VAE-c16   & $c\!=\!16$  & --- & --- & --- & --- & --- & --- & 0.66 & \textbf{0.11} \\
\midrule
VAE-1e-5  & $\lambda_\text{KL}\!=\!10^{-5}$ & --- & --- & --- & --- & --- & --- & 0.20 & \textbf{0.14} \\
VAE-1e-6  & $\lambda_\text{KL}\!=\!10^{-6}$ & --- & --- & --- & --- & --- & --- & 0.41 & \textbf{0.13} \\
VAE-1e-7  & $\lambda_\text{KL}\!=\!10^{-7}$ & --- & --- & --- & --- & --- & --- & \textbf{0.09} & 0.10 \\
AE    & $\lambda_\text{KL}\!=\!0$       & --- & --- & --- & --- & --- & --- & 0.09 & \textbf{0.07} \\
\bottomrule
\end{tabular}
\end{table}

We computed a descriptive fixed-effect decomposition of the corrected sum of squares of $\log(\mathrm{FID\mbox{-}192})$ over the balanced 3$\times$3$\times$6
default discrete panel (Table~\ref{tab:factorial_effects}), with quantizer
family, vocabulary size and generator family as factors and each fraction
given by the corrected sum of squares for that term divided by the total
corrected sum of squares. For this complete balanced design the orthogonal
effect decomposition is order-invariant. Because each cell is trained once, the decomposition is descriptive, with no seed-variance term or $p$-values and a residual that is not experimental noise. Generator family accounts
for the largest share of this variation (59.5\%) and quantizer family for
about a sixth of that (10.1\%), and the quantizer--generator and vocabulary--generator interactions
together account for about a quarter (23.4\%). Tokenizers and generators should therefore be evaluated jointly. We rely only on the qualitative presence of the quantizer--generator interaction, which the three-seed vocabulary-1024 block corroborates (Table~\ref{tab:seed_variance}). Generator families also
differ in parameter count, optimiser settings and training duration, so the
decomposition summarises differences between training recipes and cannot
attribute variance to architecture alone.

\begin{table}[h]
\centering
\caption{Descriptive fixed-effect decomposition over the 54 default
discrete cells, using $\log(\mathrm{FID\mbox{-}192})$ as the response.
The panel is a balanced VQ/LFQ/FSQ $\times$ 1K/2K/4K $\times$ six-generator
grid with positive FID values. Fractions are corrected sums of squares
divided by the total corrected sum of squares (descriptive, single-seed).}
\label{tab:factorial_effects}
\small
\begin{tabular}{@{}lrl@{}}
\toprule
\textbf{Source} & \textbf{Corrected SS fraction} & \textbf{Interpretation} \\
\midrule
Generator family & 59.5\% & largest main effect \\
Vocabulary $\times$ generator & 15.0\% & larger vocabularies help some generators, hurt others \\
Quantizer family & 10.1\% & LFQ default advantage \\
Quantizer $\times$ generator & 8.4\% & generator rankings depend on tokenizer family \\
Vocabulary & 4.4\% & smaller main effect than the interactions \\
Other interactions & 2.5\% & quantizer--vocabulary and three-way residual \\
\bottomrule
\end{tabular}
\end{table}

\paragraph{Sampling hyperparameters narrow the gap to the AR baseline, depending on the tokenizer.}
The values in Table~\ref{tab:main_results} use each generator's default
sampling configuration (temperature 0.9, full step budget). To test
sensitivity to this choice, we ran sampling-hyperparameter sweeps on all
three vocabulary-1{,}024 tokenizers, with per-generator grids over
temperature, step count, $k$ and (where applicable) mask schedule screened on
2K samples of the official MedMNIST validation split ($n\!=\!11{,}219$) and
each selected configuration then evaluated once on the test split at 10K
samples. The full per-generator screening grids are listed in
App.~\ref{app:implementation}. Per tokenizer the grids contain 60 MaskGIT, 30 BFN, 25 DFM, 20 D3PM, 20 SEDD and 18 AR configurations, so the large D3PM and SEDD gains did not come from a larger search.
Table~\ref{tab:hp_sweep} reports these validation-selected tuned FIDs alongside
the default-configuration FIDs for the 18 (tokenizer, generator) cells.
Tuning improves every generator on at least one tokenizer, by an amount that depends on the tokenizer.

\begin{table}[h]
\centering
\caption{Sampling-hyperparameter sweep at vocabulary 1{,}024 (FID-192 $\downarrow$). Each cell shows default FID $\rightarrow$ tuned FID, both at 10K test samples. Hyperparameters are selected on a held-out validation split (2K validation screen, $n\!=\!11{,}219$) and the selected configuration is then evaluated once on the test split at 10K samples; the 2K validation screen noise floor (real-vs-real FID-192) is 0.010. Bold marks tuned values that improve over the default.}
\label{tab:hp_sweep}
\small
\begin{tabular}{@{}lccc@{}}
\toprule
\textbf{Generator} & \textbf{LFQ-1024} & \textbf{FSQ-1024} & \textbf{VQ-1024} \\
\midrule
AR Transformer & $0.33 \rightarrow 0.33$ & $0.39 \rightarrow \textbf{0.29}$ & $0.52 \rightarrow \textbf{0.46}$ \\
MaskGIT        & $1.91 \rightarrow \textbf{1.89}$ & $1.15 \rightarrow \textbf{0.88}$ & $1.86 \rightarrow \textbf{1.42}$ \\
DFM            & $1.77 \rightarrow \textbf{1.58}$ & $3.02 \rightarrow \textbf{1.47}$ & $4.42 \rightarrow \textbf{3.13}$ \\
D3PM           & $0.44 \rightarrow \textbf{0.09}$ & $1.21 \rightarrow \textbf{0.13}$ & $1.30 \rightarrow \textbf{0.15}$ \\
SEDD        & $0.41 \rightarrow \textbf{0.10}$ & $1.05 \rightarrow \textbf{0.10}$ & $1.31 \rightarrow \textbf{0.13}$ \\
BFN            & $2.27 \rightarrow \textbf{2.13}$ & $7.60 \rightarrow \textbf{5.80}$ & $5.98 \rightarrow \textbf{4.10}$ \\
\bottomrule
\end{tabular}
\end{table}

Both discrete-diffusion generators are sensitive to step-count retuning. On all three tokenizers D3PM and SEDD drop from
1{,}000 steps to 100--500 (and on LFQ-1024 the selected temperature
also falls from $T\!=\!0.9$ to $T\!=\!0.7$, whereas FSQ-1024 retains
$T\!=\!1.0$), yielding $4$--$10\times$ FID improvements and measured inference
speed-ups; on VQ-1024, D3PM falls from 1.30 to 0.15 and SEDD
from 1.31 to 0.13. With these tuned settings, D3PM on LFQ-1024, FSQ-1024
and VQ-1024 (0.09, 0.13 and 0.15) overtakes the corresponding AR baselines,
inverting the default-configuration ranking. At this
token-grid size, iterations beyond $\sim$100--500 steps appear to add
sampling noise without improving quality. The reduced-step sampler computes a new
$N$-step cosine absorbing schedule for the trained checkpoint and does not
respace the original 1{,}000-step chain with arbitrary jumps
(App.~\ref{app:implementation}).

To test whether the discrete-diffusion gains are specific to one training
seed, the same validation-selected tuning protocol was re-run on the two
retrained seeds (43 and 44) for the two most sensitive LFQ-1024 cells
(Table~\ref{tab:tuned_seeds}). The selected-configuration test FID is at or below 0.15 at every seed for both
generators, with a mean of 0.11 that is about four times lower than the
defaults of 0.44 and 0.41, so the tuned operating point
is stable across seeds. The selected setting itself does not transfer, since applying the seed-42 D3PM configuration unchanged to the seed-43 model gives 0.34.

\begin{table}[h]
\centering
\caption{Reduced-step tuned FID-192 ($\downarrow$, 10K test samples) at three
training seeds, for the two most sampler-sensitive LFQ-1024 cells. Each seed
is tuned independently by the same held-out validation protocol as
Table~\ref{tab:hp_sweep}. The default (untuned) FID is shown for reference.}
\label{tab:tuned_seeds}
\small
\begin{tabular}{@{}lcccc c@{}}
\toprule
\textbf{Cell} & \textbf{Default} & \textbf{Seed 42} & \textbf{Seed 43} & \textbf{Seed 44} & \textbf{Mean} \\
\midrule
LFQ-1024 + D3PM    & 0.44 & 0.09 & 0.15 & 0.10 & 0.11 \\
LFQ-1024 + SEDD & 0.41 & 0.10 & 0.11 & 0.13 & 0.11 \\
\bottomrule
\end{tabular}
\end{table}

MaskGIT and BFN gain less from tuning, and least on LFQ. MaskGIT's gains
grow from negligible on LFQ ($-1\%$) to $-23\%$ and $-24\%$ on FSQ and VQ, and
BFN's grow from $-6\%$ on LFQ to $-24\%$ and $-31\%$. Both generators are
confidence-driven samplers (MaskGIT's iterative unmasking, BFN's
Bayesian update) that may benefit more from temperature/step retuning
when their inputs come from less uniform token distributions. Even with
this benefit BFN's best tuned cell remains 2.13
(LFQ-1024, $T\!=\!1.2$, 100 steps); an additional sweep on BFN's best
default cell (LFQ-2048, default 1.46) plateaus at 1.38 across 30
configurations, suggesting that BFN's gap is not fully explained by the
sampling grid we tested.

Retuning improves DFM moderately on all
three tokenizers (1.77 to 1.58, 3.02 to 1.47 and 4.42 to 3.13 on LFQ, FSQ and
VQ), with LFQ and FSQ selecting 200 Euler steps, but it remains behind the
tuned D3PM and SEDD cells.

\begin{figure}[t]
\centering
\includegraphics[width=\linewidth]{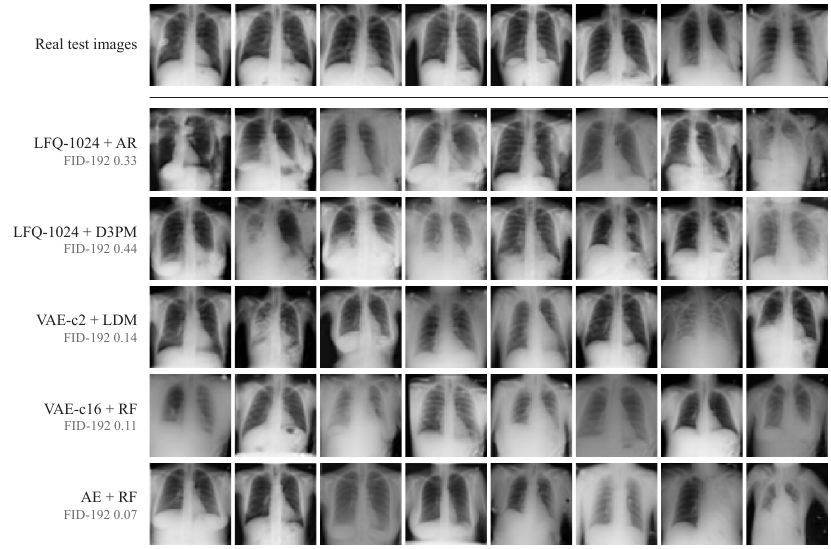}
\caption{Uncurated samples from five of the lowest-FID default cells, with real test images in the top row. All samples use default sampling and a fixed sampling seed.}
\label{fig:sample_grid}
\end{figure}

\begin{figure}[t]
\centering
\includegraphics[width=\linewidth]{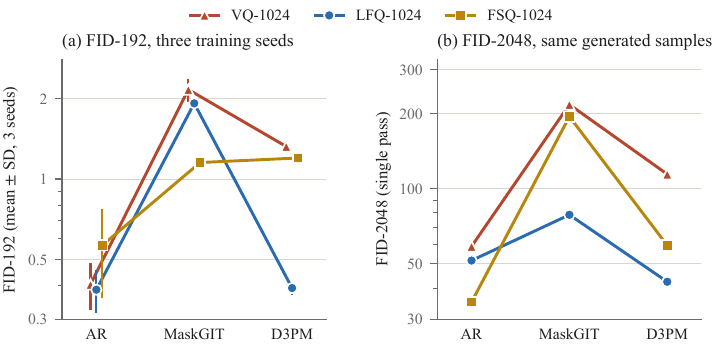}
\caption{The vocabulary-1024 block as an interaction plot. (a) FID-192, mean and standard deviation over three training seeds; points are offset horizontally for legibility. (b) Standard FID-2048 computed on the same generated samples, one generation per cell. Both axes are logarithmic and lower is better. If quantizer and generator effects were separable the three lines would be parallel.}
\label{fig:interaction}
\end{figure}

\paragraph{Across the matched-vocabulary grid, the best quantizer changes with the generator.}
Holding vocabulary constant at each of 1{,}024, 2{,}048 and 4{,}096 separates
quantization choice from vocabulary size. Across the single-seed
grid LFQ gives the lowest FID in 15 of 18 cells, with FSQ taking two
(MaskGIT at vocabularies 1{,}024 and 2{,}048) and VQ one (AR at
4{,}096), but the ordering is generator-dependent and several gaps
are small, so this count is treated as descriptive only. To test the
interaction against training-seed variance, the full vocabulary-1024 block,
$\{$VQ, LFQ, FSQ$\}\times\{$AR, MaskGIT, D3PM$\}$ plus LFQ+SEDD, was retrained
at three training seeds (42/43/44), with all seeds of a quantizer evaluated on
one GPU (Table~\ref{tab:seed_variance} and Figure~\ref{fig:interaction}). The median per-cell standard
deviation is 0.047, ranging from 0.012 (FSQ+MaskGIT) to 0.214 (VQ+MaskGIT), with
the D3PM and SEDD cells between 0.022 and 0.043. Of the nine pairwise
quantizer comparisons in this block, four exceed three pooled seed standard
deviations, and these four carry the interaction, in which the best quantizer
changes with the generator. Under D3PM LFQ beats both FSQ and VQ, whereas
under MaskGIT FSQ beats both LFQ (by 0.77, well above its 3-SD threshold)
and VQ. Standard FID-2048 on the same samples (Table~\ref{tab:fid_dual})
keeps the D3PM ordering and also has the best quantizer changing with the
generator, but it prefers LFQ under MaskGIT and FSQ under AR. The seeded
block therefore supports the interaction itself and the D3PM ordering; the AR and MaskGIT winners depend on the metric and are left unresolved.

\begin{table}[h]
\centering
\caption{Training-seed variance of default-config FID-192 ($\downarrow$, 10K
test samples) over the vocabulary-1024 interaction block, at three seeds. Seed
42 is the main-table run; seeds 43/44 are retrained from scratch with the
frozen tokenizer. VQ seed 42 is the main-grid cell and all VQ seeds
were trained and evaluated on the main-grid A6000. LFQ and FSQ rows were evaluated on
an L40S; re-evaluating six seed-42 cells on both GPUs gives a mean difference of
$-0.004$. Small differences from Table~\ref{tab:main_results} (e.g.\ LFQ+AR 0.34
here vs.\ 0.33 there) are recomputation drift within the estimator noise
floor.}
\label{tab:seed_variance}
\small
\begin{tabular}{@{}lccc cc@{}}
\toprule
\textbf{Cell} & \textbf{Seed 42} & \textbf{Seed 43} & \textbf{Seed 44} & \textbf{Mean} & \textbf{Std} \\
\midrule
VQ-1024 + AR       & 0.52 & 0.35 & 0.35 & 0.40 & 0.080 \\
VQ-1024 + MaskGIT  & 1.86 & 2.36 & 2.26 & 2.16 & 0.214 \\
VQ-1024 + D3PM     & 1.30 & 1.38 & 1.29 & 1.32 & 0.043 \\
\midrule
LFQ-1024 + AR      & 0.34 & 0.34 & 0.49 & 0.39 & 0.070 \\
LFQ-1024 + MaskGIT & 1.90 & 1.86 & 1.99 & 1.92 & 0.052 \\
LFQ-1024 + D3PM    & 0.42 & 0.40 & 0.36 & 0.39 & 0.022 \\
LFQ-1024 + SEDD & 0.41 & 0.42 & 0.48 & 0.44 & 0.032 \\
\midrule
FSQ-1024 + AR      & 0.40 & 0.86 & 0.44 & 0.57 & 0.207 \\
FSQ-1024 + MaskGIT & 1.15 & 1.14 & 1.17 & 1.15 & 0.012 \\
FSQ-1024 + D3PM    & 1.19 & 1.15 & 1.25 & 1.20 & 0.042 \\
\bottomrule
\end{tabular}
\end{table}

The five comparisons that fall within seed noise are left unranked: all
three under AR, where the quantizers are level (for example LFQ against FSQ,
mean gap 0.18 against a 3-SD threshold of 0.65), VQ against LFQ under
MaskGIT, and VQ against FSQ under D3PM (gap 0.12 against 0.18). All other cells, including DFM, BFN and the VQ and FSQ SEDD cells at
vocabulary 1{,}024 and every cell at 2{,}048 and 4{,}096, remain single-seed,
so the seed-verified interaction applies only to the seeded block. FSQ's
generator-specific failure modes, most clearly on D3PM/SEDD/BFN at vocabulary
4{,}096, and VQ's competitiveness under AR but not under the iterative samplers both indicate that vocabulary
size, quantizer family and generator family are joint design choices.

\paragraph{VQ recovers at appropriate codebook sizes, mainly with AR.}
VQ with 8{,}192 codes gave catastrophic generation quality in pilot
experiments (FID $\gg 100$ across all generators). At 1{,}024--4{,}096 codes
VQ produces competitive generation FIDs, and with AR it is competitive with the
best lookup-free cells (0.52, 0.52 and 0.31 at the three vocabularies, against
0.33--0.45 for LFQ and 0.37--0.73 for FSQ). VQ-4096 + AR is the lowest single
default discrete cell, but its margin over LFQ-1024 + AR (0.33) and FSQ-4096 +
AR (0.37) is within the seed spread of the AR cells
(Table~\ref{tab:seed_variance}), so it is not ranked above them. Under the
five iterative generators VQ trails LFQ at every vocabulary except under
MaskGIT at 1{,}024 and 2{,}048, where the two are level. VQ is therefore usable at these vocabulary sizes, but its competitiveness depends on the generator.

\paragraph{LFQ's high marginal entropy co-occurs with its default advantage but does not explain it.}
LFQ's near-uniform marginal token distribution
(Table~\ref{tab:tokenizer_props}) co-occurs with better default FID in most
matched cells, but marginal entropy does not order the tokenizers by
generation quality (Spearman $-0.20$ against best FID-192;
\S\ref{sec:predictor}). FSQ's two wins are both under MaskGIT's iterative
unmasking (vocabularies 1{,}024 and 2{,}048), where the fixed grid's
structured codes may be easier to commit to incrementally. All discrete
generators consume flattened categorical token IDs through learned embedding
and softmax layers; we do not test geometry-aware bit heads for LFQ or
factorised scalar heads for FSQ. We did not observe codebook collapse, the VQ failure mode that motivated LFQ in MAGVIT-v2, at any of the three vocabularies tested.

\subsection{The Rate-Distortion-Modelability Framing}
\label{sec:tradeoff}

\begin{figure}[h]
\centering
\includegraphics[width=\linewidth]{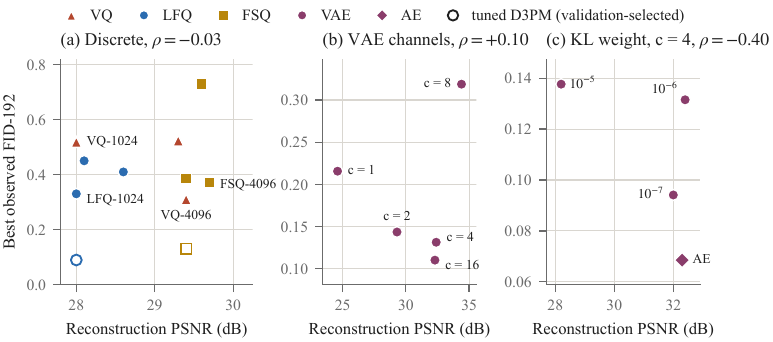}
\caption{Reconstruction PSNR against the lowest default-configuration FID-192 over all generators paired with each tokenizer, with Spearman $\rho$ in each panel title. (a) The nine discrete tokenizers; hollow markers are the validation-selected D3PM results for LFQ-1024 and FSQ-1024. (b) VAE channel count $c$ at $\lambda_\text{KL}=10^{-6}$. (c) KL weight at $c=4$, where AE is the $\lambda_\text{KL}=0$ endpoint; with four points this panel is descriptive only.}
\label{fig:recon_vs_gen}
\end{figure}

When each tokenizer is paired with its best observed generator,
reconstruction quality (PSNR) does not reliably predict generation
quality (best FID across available generators; Fig.~\ref{fig:recon_vs_gen}).
Spearman correlations are near zero in the discrete and VAE-channel
panels ($\rho\!=\!-0.03$ and $+0.10$), and the fixed-channel KL panel is
treated as descriptive only. This agrees
with recent natural-image evidence that reconstruction metrics alone are
insufficient for selecting tokenizers~\citep{yao2025vavae,xiong2025gigatok}.
The best- and worst-reconstruction discrete tokenizers
(FSQ-4096 PSNR 29.7 vs LFQ-1024 PSNR 28.0, the lowest discrete PSNR alongside VQ-1024) produce comparable
best-generation FIDs (0.37 vs 0.33), and the lowest best-generation FID among
the discrete tokenizers belongs to VQ-4096 (0.31) at a mid-range 29.4 dB. The same pattern appears in the
continuous panel, where VAE-c8 is the best reconstructor (PSNR 34.4) yet
produces the worst generation in the channel sweep, while AE (PSNR 32.3)
yields the lowest generation FID anywhere in Table~\ref{tab:main_results}
(RF, 0.07).

This pattern can be accounted for by a rate-distortion-modelability framing.
At $64 \times 64$ resolution with 64 tokens, larger vocabularies provide
diminishing reconstruction gains whilst increasing the categorical state
space and potentially the generator's prediction burden at each token
position, so the operating point must balance reconstruction fidelity against
the generator's ability to model the resulting token distribution. This is the
\emph{rate-distortion-modelability} tradeoff discussed by
\citet{dieleman2025rdm}, and our evidence supports it as an interpretation of this controlled setting but does not establish a universal threshold for medical images.

\paragraph{Discrete vs.\ continuous generation.}
At default sampling configurations the best continuous cells beat every discrete cell, with VAE-c2 + LDM (FID 0.14) and VAE-c16 + RF (FID 0.11) both
outperform LFQ-1024 + AR (0.33), the best discrete result, although the
capacity trend on the continuous side is non-monotonic. RF improves from
$c\!=\!1$ (0.26) to $c\!=\!4$ (0.13), regresses at $c\!=\!8$ (0.32)
and recovers at $c\!=\!16$ (0.11), whilst LDM's best is $c\!=\!2$ (0.14). The continuous generators were trained for 200 epochs and the
discrete generators for 100, so this default comparison also reflects the
reported recipe and training budget in addition to latent type. Generation therefore does not improve monotonically with channel count, which fits the rate-distortion-modelability framing better than the rate-distortion frontier alone. The HP sweep on the discrete side (Table~\ref{tab:hp_sweep}) almost
closes the discrete--continuous gap, with tuned LFQ-1024 + D3PM reaching FID
0.09, comparable under FID-192 to the lowest-FID default LDM cells
(VAE-$10^{-7}$ and AE, both 0.09) and trailing only AE + RF (0.07; see
Fig.~\ref{fig:sample_grid}). This compares a tuned operating point against default continuous samplers, which were not given an equivalent sweep. Several of the lowest continuous results involve weakly
regularised or unregularised latents, although VAE-$10^{-7}$ remains
approximately unit-scale. The remaining advantage of continuous tokenizers may partly reflect the choice of sampling budget as well
as the latent representation.

\subsection{Generator-Free Statistics Do Not Predict Modelability}
\label{sec:predictor}

Since reconstruction does not identify the best tokenizer, we tested whether any generator-free property of the token field predicts downstream generation quality. For each discrete tokenizer we computed marginal token entropy, per-position entropy dispersion and a \emph{neighbour-conditional predictive gain} directly on the
tokenized data, with the last defined as the reduction in a token's entropy
given its left or upper neighbour, estimated from bigram counts fitted on the
training split and scored as held-out cross-entropy on the test split. The
smoothing constant is selected on a validation split and never against FID.

\begin{table}[h]
\centering
\caption{Generator-free statistics against the best default FID-192 of each
discrete tokenizer ($n=9$, Spearman $\rho$).}
\label{tab:predictor}
\small
\begin{tabular}{@{}lc@{}}
\toprule
\textbf{Statistic} & \textbf{Spearman $\rho$ with best FID-192} \\
\midrule
Neighbour-conditional predictive gain, left & $+0.05$ \\
Neighbour-conditional predictive gain, up & $+0.17$ \\
Per-position entropy dispersion & $+0.03$ \\
Reconstruction PSNR & $-0.03$ \\
Marginal token entropy & $-0.20$ \\
\bottomrule
\end{tabular}
\end{table}

None of these statistics tracks generation quality
(Table~\ref{tab:predictor}). Predictive gain does separate the three VQ
tokenizers from the six LFQ and FSQ ones, but VQ is not the
weaker family by best FID, so the separation is uninformative about modelability; it also holds only for some bigram smoothing settings.

\subsection{Generator Architecture Analysis}
\label{sec:generators_results}

\paragraph{Autoregressive vs.\ iterative discrete generators.}
With default sampling, the AR transformer wins all nine discrete rows in
Table~\ref{tab:main_results}. Its margin over the best iterative generator is
largest on VQ, where the next-best default cell is 2.5 to 6.5 times worse,
and smallest on LFQ-1024 and LFQ-2048 (1.2 and 1.6 times). The HP sweep on LFQ-1024 (Table~\ref{tab:hp_sweep})
qualifies this picture: with tuned step counts and temperatures, D3PM
(0.09) and SEDD (0.10) overtake AR (0.33), so AR's default-configuration
advantage is partly attributable to the iterative samplers being run at their
default 1{,}000-step settings instead of their validation-selected operating points. Across the three seeds this default
LFQ-1024 + D3PM cell has mean FID 0.39 (std 0.022;
Table~\ref{tab:seed_variance}), so the seed spread is far smaller than the
sampler-retuning shift from 0.44 to 0.09 (see also Table~\ref{tab:tuned_seeds}).

\paragraph{SEDD and D3PM differ little at default configurations.}
SEDD and D3PM share the same architecture, forward process and sampler, and
differ in the training objective, which in this implementation is a hybrid
score-entropy objective for SEDD and the variational bound for D3PM. Across
all nine discrete cells the two are within 0.1 FID of each other on most rows (mean absolute gap 0.10); SEDD wins five cells and D3PM four, and five of the nine gaps are at most 0.06 FID and likely within sampling variance. The
training objective alone is therefore a small factor in these
default-configuration recipes compared with vocabulary, tokenizer or sampling
configuration, although this result does not extend to canonical SEDD with
tau-leaping. The validation-selected LFQ-1024 configurations differ,
with D3PM best at $T\!=\!0.7$ and 100 steps (FID 0.090) and SEDD best at
$T\!=\!0.7$ and 500 steps (FID 0.098), so SEDD's best configuration used a
$5 \times$ larger step budget than D3PM's at matched temperature.

\paragraph{Latent diffusion vs.\ rectified flow.}
Both continuous generators share the same DiT backbone and operate on
identical VAE latents within each tokenizer row, and differ in training
target, sampler and default step budget, with a stochastic differential
equation (SDE) sampler for LDM and a straight-line ordinary differential
equation (ODE) sampler for RF.
Across the channel sweep, the two columns disagree
on which capacity is optimal: LDM minimises at $c\!=\!2$ (0.14) and
degrades at higher capacities (c4 0.41, c8 1.61, c16 0.66), whereas RF
improves through $c\!=\!4$ (0.13), spikes at $c\!=\!8$ (0.32), and
reaches its global minimum at $c\!=\!16$ (0.11). The c16 LDM--RF discrepancy is plausibly explained
mechanistically by the latent statistics, since VAE-c16's encoded latents
have $\mathrm{std}\!=\!0.62$ and $|z|_{\text{mean}}\!=\!0.35$ on the test
split, smaller than c1--c8 (all $\mathrm{std} \in [0.91, 1.02]$), which
suggests that the c16 model partially collapsed its latent magnitudes during
training. LDM's fixed-variance
noise schedule is sensitive to latent scale because the scale changes the
effective signal-to-noise schedule, while RF's straight-line ODE appears
less sensitive. A uniform latent-normalisation
ablation is needed before attributing this fully to the RF objective. The c8
spike in both columns is harder to explain, as c8's latent statistics
($\mathrm{std}\!=\!0.93$, $|z|_{\text{mean}}\!=\!0.72$) are within the
c1--c4 range and the latent-magnitude explanation for LDM does not apply. The
most likely candidates are a training-stage artifact (a single seed at this
specific capacity reaching a poor local minimum) or an interaction
between the c8 latent capacity and this dataset, and separating
these would require multi-seed retraining, which is left to future work. RF
is also about $10\times$ faster (Table~\ref{tab:throughput}: 37.7 vs 3.63
samples/s; 100 ODE steps vs LDM's 1{,}000 DDPM steps) at matched or better
quality on every continuous row except $c\!=\!1$, $c\!=\!2$ and
$\lambda_\text{KL}\!=\!10^{-7}$, so at MedMNIST scale RF is the better
speed--quality default of the two continuous-latent generators. RF also has the
smaller range across the channel sweep ($\max - \min$ FID 0.21
against 1.47 for LDM), with LDM's range driven mainly by its $c\!=\!8$
outlier, with no smooth dependence on channel count. LDM's main remaining
advantage is compatibility with standard latent-diffusion practice; its
range across the sweep is larger, not smaller.

\paragraph{Inference throughput.}
Table~\ref{tab:throughput} reports generation throughput for each architecture. All models use matched transformer backbones, but parameter counts differ between AR/MaskGIT and diffusion-style variants; throughput is governed mainly by the number of full-model forward passes, with a smaller per-step-cost effect that separates only same-step-count generators (for example BFN versus D3PM and SEDD at 1{,}000 steps).

\begin{table}[h]
\centering
\caption{Generation throughput for 10K samples at $8\times8$ token resolution and vocabulary 1{,}024 on one NVIDIA RTX A6000, the GPU model of the main grid. Timings cover sampling and decoding but not model loading or metric computation. Samples/s is the median of three timed passes after a warm-up batch (one pass for the 1{,}000-step generators), in float32 at batch size 128 (64 for the tuned rows), with randomly initialised weights, since the cost of a step does not depend on them. Default rows use the canonical settings; tuned rows are the validation-selected LFQ-1024 settings. ``Parallel decoding'' indicates whether several tokens are predicted per forward pass.
}
\label{tab:throughput}
\small
\begin{tabular}{@{}lrrrr@{}}
\toprule
Generator & Steps & Samples/s & Time for 10K & Parallel decoding \\
\midrule
MaskGIT                       & 12   & 333  & 30\,s    & yes \\
AR transformer                & 64   & 146  & 68\,s    & no \\
DFM                           & 100  & 37.0 & 270\,s   & yes \\
D3PM                          & 1000 & 3.67 & 2726\,s  & yes \\
D3PM, tuned (LFQ-1024)        & 100  & 33.2 & 301\,s   & yes \\
SEDD                          & 1000 & 3.67 & 2724\,s  & yes \\
SEDD, tuned (LFQ-1024)        & 500  & 6.60 & 1515\,s  & yes \\
BFN                           & 1000 & 3.48 & 2878\,s  & yes \\
RF                            & 100  & 37.7 & 265\,s   & yes \\
LDM                           & 1000 & 3.63 & 2755\,s  & yes \\
\bottomrule
\end{tabular}

\end{table}

MaskGIT's 12 bidirectional passes are the cheapest route to a sample (about 330 samples/s), followed by the AR transformer's 64 key-value-cached passes (about 150 samples/s); the 100-step DFM and RF reach about 37 samples/s, and the 1{,}000-step generators run at 3.5--3.7 samples/s. Retuning LFQ-1024 + D3PM to 100 steps raises its throughput from 3.7 to 33 samples/s, close to DFM and RF, and the selected 500-step SEDD setting raises SEDD's from 3.7 to 6.6 samples/s. These tuned settings narrow the wall-clock gap to AR, with tuned D3PM about 4.4 times slower than AR, but do not remove it.

\section{Extensible 2D/3D Libraries}
\label{sec:libraries}

All experiments were run with two libraries.
\medtok{} provides a common encoder--decoder for 2D and 3D
medical-image tokenization with interchangeable VQ, LFQ, FSQ, Residual
FSQ, VAE and AE quantization heads, so a new quantizer only replaces
the head. \medlat{} provides the
corresponding latent generators (AR transformers, MaskGIT, DFM, D3PM,
SEDD, BFN, LDM and RF) on shared DiT-style backbones. Both libraries
accept 2D images and 3D volumes, so a new tokenizer can be paired
with every existing generator and a new generator can be run against
the existing tokenizers. This allows the
tokenizer--generator--sampler comparison reported here to be repeated on
datasets other than ChestMNIST.

\section{Discussion}
\label{sec:discussion}

\subsection{What the Controlled Grid Implies for Practice}

\paragraph{Joint evaluation.}
Tokenizer selection for
low-resolution medical-style image generation should be evaluated jointly
with generator architecture and sampling configuration. Under default
sampling, LFQ-1024 is the best discrete choice in our
matched-vocabulary grid and AE/RF gives the lowest
continuous-latent FID-192, although the tuned D3PM/SEDD results show that
default sampling budgets can mis-rank iterative discrete generators.

\paragraph{Reconstruction alone is insufficient for tokenizer selection.}
As shown in \S\ref{sec:tradeoff}, reconstruction PSNR does not reliably
predict generation FID on either the discrete vocabulary sweep or the
continuous channel sweep, although this is not evidence for a universal
reconstruction--generation tradeoff. A
reconstruction-FID decomposition in the same feature spaces used for
generation would test this directly.

\paragraph{Choosing a quantizer.}
Recent literature has favoured lookup-free methods (LFQ, FSQ) over VQ, citing
codebook collapse and poor scaling, and the controlled $3 \times 3$ comparison
here partially supports this preference. LFQ gives the
lowest FID in 15 of 18 cells of the single-seed grid, and in the
seed-verified vocabulary-1024 block the best quantizer changes with the
generator (the D3PM ordering LFQ $<$ FSQ $<$ VQ holds under both FID-192 and
FID-2048, while the AR and MaskGIT winners depend on the metric). VQ at
appropriately sized vocabularies avoids catastrophic failure and is
competitive with AR (best VQ cell: VQ-4096 + AR, 0.31, level with the best LFQ
default cell at 0.33), but trails LFQ under most iterative generators, so the
advantage of lookup-free methods in this grid depends on the generator.
Practitioners targeting a
new medical-image domain should therefore treat vocabulary size and quantizer family
as coupled tuning axes, with LFQ as the strongest default in our grid and
FSQ/VQ as plausible alternatives for specific generator families.

\paragraph{Sampling-budget tuning matters for absorbing-state diffusion on every tokenizer.}
The sampling-hyperparameter sweep across all three tokenizers at vocabulary 1{,}024
(Table~\ref{tab:hp_sweep}) shows that the conventional 1{,}000-step
defaults were suboptimal for D3PM and SEDD on every tokenizer; reducing the
step budget to 100--500 and adjusting the temperature reduces FID by
$4$--$10\times$ whilst cutting sampling cost in proportion to the step
count. The selected budget differs by tokenizer (D3PM selects 100, 250 and
500 steps on LFQ, FSQ and VQ). Comparisons of discrete diffusion variants in
prior literature typically use the canonical 1{,}000-step setting and may
therefore understate these methods' best operating points; the same caveat
applies to our own default-configuration results in
Table~\ref{tab:main_results}.

\paragraph{Sampler tuning can reorder tokenizers.}
Because the tuning gains differ by generator and tokenizer, tuning also
changes tokenizer comparisons. At default settings D3PM separates the three
tokenizers widely (0.44, 1.21 and 1.30 for LFQ, FSQ and VQ), whereas after
tuning they lie within 0.06 of each other (0.09, 0.13 and 0.15), and MaskGIT
and BFN gain least on LFQ. These tuned values are single-seed, but they
indicate that the appropriate experimental unit is the (tokenizer, generator,
sampling-config) triple, with (tokenizer, generator) pairs compared only
under matched sampling. BFN's best swept cell remains far above
the diffusion-style floor (FID 2.13; \S\ref{sec:vocab}).

\begin{figure}[h]
\centering
\includegraphics[width=\linewidth]{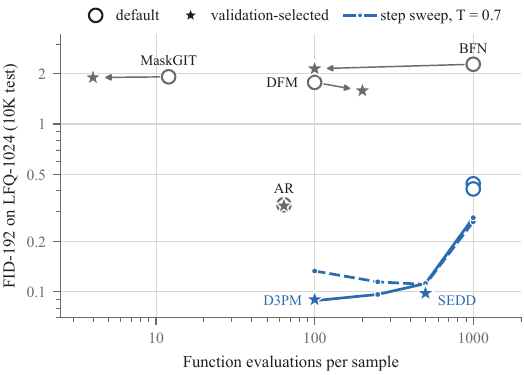}
\caption{Sampling budget against FID-192 on LFQ-1024, all at 10K test samples. Open circles are the default configurations and stars the validation-selected ones; arrows join the two where both the budget and the score change. Blue lines are D3PM (solid) and SEDD (dashed) evaluated at 100, 250, 500 and 1000 steps with temperature 0.7. The SEDD line and the selected SEDD star are separate generations of the same 500-step setting (0.111 and 0.098).}
\label{fig:nfe_vs_fid}
\end{figure}

\paragraph{Inference budget vs.\ generation quality.}
Figure~\ref{fig:nfe_vs_fid} synthesises the inference-time cost of each
generator family as the best observed FID at each nominal number of function
evaluations (NFE) per sample, sweeping the
NFE-controlling hyperparameter (timesteps for D3PM, SEDD and BFN,
refinement steps for MaskGIT, Euler steps for DFM) and
taking the best FID over the swept temperatures at each NFE on LFQ-1024.
D3PM and SEDD have steep, U-shaped curves
with best points at $\sim$100--500 NFE, and increasing the step
count above 500 worsens FID, contrary to the assumption behind their
canonical 1{,}000-step defaults (open circles). MaskGIT and DFM change
little with NFE (DFM's best validation FID falls from 1.93 at 10 Euler steps
to 1.56 at 200), consistent with their low-step design, and their best operating points are within a factor of
two of their canonical defaults. BFN sits above all other discrete
generators across the NFE range, with no NFE setting bringing it within
$5\times$ of the tuned D3PM/SEDD floor, and so the tested sampling budgets
do not appear to account for the whole gap. The AR transformer (horizontal
reference; fixed NFE equal to the sequence length of 64, FID 0.33) is competitive
only against the tuned iterative generators. For these 64-token grids, absorbing-state
diffusion works best well below its canonical 1{,}000-step setting (D3PM is best
near 100 NFE and SEDD near 500), so tuned D3PM narrows the nominal NFE gap to
AR. Measured wall-clock comparisons still favour cached AR over the tuned
discrete-diffusion settings (Table~\ref{tab:throughput}).

\subsection{Robustness Checks and Downstream Demonstrations}

\paragraph{Cross-dataset replication: PneumoniaMNIST and OrganAMNIST.}
To probe whether the ChestMNIST findings generalise within the MedMNIST
family, we retrained the six representative cells on two additional
datasets at $64 \times 64$, both 1-channel: PneumoniaMNIST (chest
X-ray; 4{,}708 train / 624 test, $\sim$17$\times$ fewer images than
ChestMNIST) and OrganAMNIST (abdominal CT slices; 34{,}561 train /
17{,}778 test, on the same order as ChestMNIST). Default-configuration
FIDs at 10K samples are reported in Table~\ref{tab:cross_dataset}. A real-vs-real
bootstrap on these test splits gives a PneumoniaMNIST full-split floor
of 0.053 (95\,\% interval $[0.025, 0.103]$ at $n=312$ per half) and an
OrganAMNIST floor of 0.0089 (95\,\% interval $[0.0027, 0.0269]$ at
$n=8{,}889$ per half), so the Pneumonia column should be read more
coarsely than the OrganAMNIST column.

The winning tokenizer depends on the dataset and is not explained by
    training-set size. On both ChestMNIST and OrganAMNIST
    (the two larger datasets), LFQ-1024 + AR is the best default-config
    discrete cell. On the much smaller PneumoniaMNIST,
    LFQ-1024 + D3PM is the best discrete cell (1.93) and
    VQ-1024 + AR beats LFQ-1024 + AR (2.09 vs 2.73), the reverse of their
    order on the two larger datasets. A natural
    hypothesis is that LFQ's high-entropy binary-hypercube targets need
    enough training data to populate their categorical statistics, so
    below some data threshold the learned VQ codebook adapts more
    efficiently. This was tested by subsampling the ChestMNIST
    training set to three sizes and retraining AR on LFQ-1024 and VQ-1024
    at each (Table~\ref{tab:data_scale}). LFQ
    beats VQ at every training-set size, including at 4{,}700 images
    (Pneumonia's scale), by a roughly constant margin of 0.13 to 0.18, with
    no narrowing at small $n$, refuting the hypothesis. The LFQ side of this
    control ran on an L40S and the VQ side on the main-grid A6000; re-evaluating six
    seed-42 cells on both GPUs gives a mean difference of $-0.004$ (range
    $-0.017$ to $+0.034$), well below that margin. The
    PneumoniaMNIST inversion is therefore not a training-set-size effect,
    although its cause remains unidentified. In practice, the
    tokenizer has to be chosen per dataset with a matched sweep,
    since neither reconstruction nor a size heuristic predicts the winner. The same
    dataset-dependence is visible for FSQ, the weakest AR tokenizer on
    Pneumonia (5.52) and OrganAMNIST (4.87) but competitive on Chest (0.39).

Among the continuous configurations, AE + RF is the most consistent:
    it wins all three columns of the continuous
    block (FID 0.07 / 1.05 / 0.94 on Chest / Pneumonia / OrganAMNIST). VAE-c2
    + LDM, by contrast, degrades on the two new datasets (4.14 /
    9.88 vs 0.14 on ChestMNIST). The c2-LDM advantage on ChestMNIST
    therefore appears specific to that dataset.

\begin{table}[h]
\centering
\caption{Cross-dataset replication on PneumoniaMNIST and OrganAMNIST (FID-192 $\downarrow$, 10K samples). Real test set sizes vary across datasets (Chest 22{,}433 / Pneumonia 624 / OrganAMNIST 17{,}778); the real-vs-real FID-192 floor is 0.053 on Pneumonia and 0.0089 on OrganAMNIST at the largest disjoint split available. Default sampling configurations throughout. Bold = best in column.}
\label{tab:cross_dataset}
\small
\begin{tabular}{@{}lccc@{}}
\toprule
\textbf{Cell} & \textbf{ChestMNIST} & \textbf{PneumoniaMNIST} & \textbf{OrganAMNIST} \\
              & (78K train)        & (4.7K train)            & (34.6K train) \\
\midrule
LFQ-1024 + AR        & 0.33          & 2.73          & 2.12 \\
FSQ-1024 + AR        & 0.39          & 5.52          & 4.87 \\
VQ-1024 + AR         & 0.52          & 2.09          & 3.84 \\
LFQ-1024 + D3PM      & 0.44          & 1.93          & 4.89 \\
VAE-c2 + LDM         & 0.14          & 4.14          & 9.88 \\
AE + RF          & \textbf{0.07} & \textbf{1.05} & \textbf{0.94} \\
\bottomrule
\end{tabular}
\end{table}

\begin{table}[h]
\centering
\caption{Training-set-size control (FID-192 $\downarrow$, 10K test samples).
AR is retrained on LFQ-1024 and VQ-1024 tokens from ChestMNIST subsampled to
three training-set sizes; the smallest ($\sim$4{,}700) matches the
scale of the PneumoniaMNIST training set. The full-size VQ row is the main-grid cell, which
this protocol reproduces exactly on the same GPU. Bold marks the lower FID at each size.}
\label{tab:data_scale}
\small
\begin{tabular}{@{}rccc@{}}
\toprule
\textbf{ChestMNIST train $n$} & \textbf{LFQ-1024 + AR} & \textbf{VQ-1024 + AR} & \textbf{Winner} \\
\midrule
4{,}700   & \textbf{0.95} & 1.11 & LFQ \\
20{,}000  & \textbf{0.37} & 0.49 & LFQ \\
78{,}468  & \textbf{0.34} & 0.52 & LFQ \\
\bottomrule
\end{tabular}
\end{table}

\begin{figure}[h]
\centering
\includegraphics[width=\linewidth]{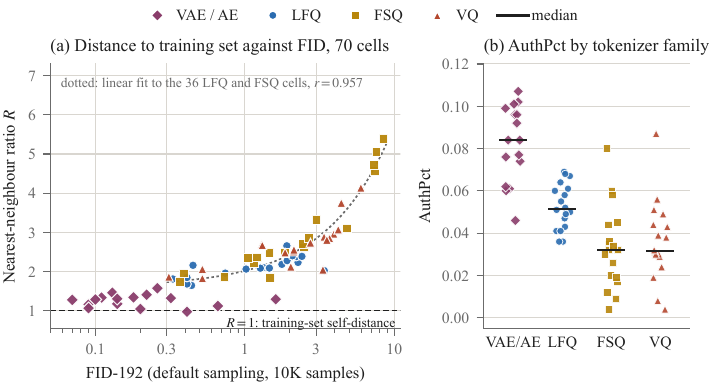}
\caption{Nearest-neighbour memorization probe in the InceptionV3 192-dimensional feature space used by FID-192, over all 70 cells. (a) $R$ is the median distance from a generated sample to its nearest training image divided by the median nearest-neighbour distance within the training set, so $R<1$ would indicate samples closer to training images than training images are to each other; the dotted curve is a linear fit to the 36 LFQ and FSQ cells. (b) AuthPct is the fraction of generated samples closer to a training image than that image is to its own nearest neighbour; horizontal bars are family medians.}
\label{fig:memorization}
\end{figure}

\paragraph{Nearest-neighbour memorization screens.}
A standing concern with generative models for medical imaging is that
low-FID samples may be near-duplicates of training images. This was
probed across all 70 evaluated cells (54 discrete cells plus 16
continuous-latent reference cells; 1{,}000 samples per cell,
the same 192-dimensional InceptionV3 feature space used for FID-192)
using two scalars: a
memorization ratio
$R = \mathrm{med}(d_{g\to t}) / \mathrm{med}(d_{t\to t})$ comparing
generated-to-training nearest-neighbour distance against the training
set's own nearest-neighbour distance, and AuthPct, the fraction of
generated samples whose nearest training neighbour is closer than that
training neighbour's own nearest training neighbour. This follows the
nearest-neighbour data-copying intuition behind synthetic-data
authenticity metrics~\citep{alaa2021faithful}. $R<1$ would indicate generations are systematically closer to
training than the training set is to itself; high AuthPct would indicate
that a meaningful fraction of generations look like a single specific
training exemplar. Across all 70 cells, this feature-space probe
does not detect a nearest-neighbour memorization signature, with minimum
$R = 0.975$ (within sampling noise of 1) and maximum AuthPct $= 0.107$.
The tokenizer families separate clearly (Fig.~\ref{fig:memorization}).

Continuous tokenizers lie closest to the training
    distribution ($R \in [1.05, 1.6]$), with VAE-c16 + LDM and
    AE + LDM achieving $R \approx 1.07$--$1.12$, i.e.\ generations
    that are close to the training set's own internal diversity. AuthPct is correspondingly highest in this
    block (median $\sim$0.077), but every value remains below the
    $\sim$0.11 ceiling, so even the closest-fit cells are not
    flagged as near-duplicates.

For the lookup-free tokenizers (LFQ, FSQ), FID and $R$ are
    linearly related (Pearson $r = +0.957$, $n=36$):
    lower FID corresponds to smaller $R$, although $R$ stays above 1 in
    every cell. The fit ($R \approx 1.61 + 0.41 \cdot \mathrm{FID}$)
    places even the best default LFQ cell (FID 0.33) at $R = 1.7$,
    above the continuous range. Quantization and decoding
    effects may contribute to the residual nearest-neighbour distance gap.

The learned-codebook family (VQ) follows the same trend, with the VQ cells lying on
    the LFQ/FSQ line (mean residual $-0.02$, range $-0.93$ to $+0.56$),
    with $R \in [1.83, 4.14]$ and AuthPct comparable to FSQ (mean $0.036$,
    against $0.033$ for FSQ and $0.053$ for LFQ). Within VQ, FID and $R$ are
    correlated ($r = 0.84$), as they are for the lookup-free families.

At the ChestMNIST scale (78K training images), the feature-space screen
therefore does not show exact-copy behaviour. This is not a formal privacy
audit, and memorization risk is dataset-size-dependent; the same probe
should be re-run on smaller or more sensitive datasets.

A separate raw-space nearest-neighbour check on seven representative
cells compares generated-to-training distances against a held-out
test-to-training baseline (1{,}000 images). LFQ-1024 + AR and LFQ-1024 +
D3PM sit above the baseline in both pixel and LPIPS distance
($R_\mathrm{pixel}=1.116/1.082$, $R_\mathrm{LPIPS}=1.289/1.227$), while
VQ-1024 + MaskGIT is farther from training
($R_\mathrm{pixel}=1.22$, $R_\mathrm{LPIPS}=2.43$). The continuous cells
are closest, with AE + LDM giving $R_\mathrm{pixel}=0.873$ and
$R_\mathrm{LPIPS}=0.821$ and AE + RF giving 0.967 and 0.995. Minimum
LPIPS distances remain non-zero (0.028 for AE + LDM versus 0.032 for the
held-out real baseline), so exact copying is not shown, although the
continuous cells are the privacy-sensitive regime for any stronger audit.
Dedicated synthetic-radiograph benchmarks such as
{CheXGenBench}~\citep{dutt2025chexgenbench} formalise this privacy axis
through patient re-identification metrics, which are complementary to the
lightweight nearest-neighbour screening reported here.

\begin{figure}[h]
\centering
\includegraphics[width=\linewidth]{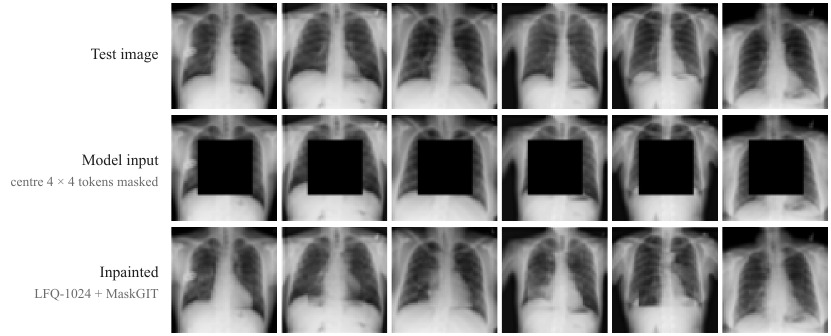}
\caption{Centre inpainting with LFQ-1024 + MaskGIT and no retraining. The central $4\times4$ block of the $8\times8$ token grid is masked and filled over 12 refinement steps.}
\label{fig:inpaint}
\end{figure}

\paragraph{Downstream demonstrations: inpainting and oracle-mask counterfactual inpainting.}
The trained generators can be used for inpainting without retraining. The discrete-diffusion / masked-LM family
(D3PM, SEDD, MaskGIT, DFM) is trained on partly-masked sequences, so
inference on a partly-known sequence is a natural operating mode, and
clamping the visible tokens at every reverse step gives conditional
generation. Figure~\ref{fig:inpaint} shows centre-mask inpainting:
the central $4\!\times\!4$ of the $8\!\times\!8$ token grid (16 of 64
tokens, corresponding to a $32\!\times\!32$ pixel block) is hidden, and
LFQ-1024 + MaskGIT fills it in over 12 iterative steps. Because this
mask is known by construction, the displayed output composites the model
prediction only inside the masked block and copies all visible pixels
from the input. The reconstructions show visually plausible
low-resolution heart and mediastinum structure.

\begin{figure}[h]
\centering
\includegraphics[width=\linewidth]{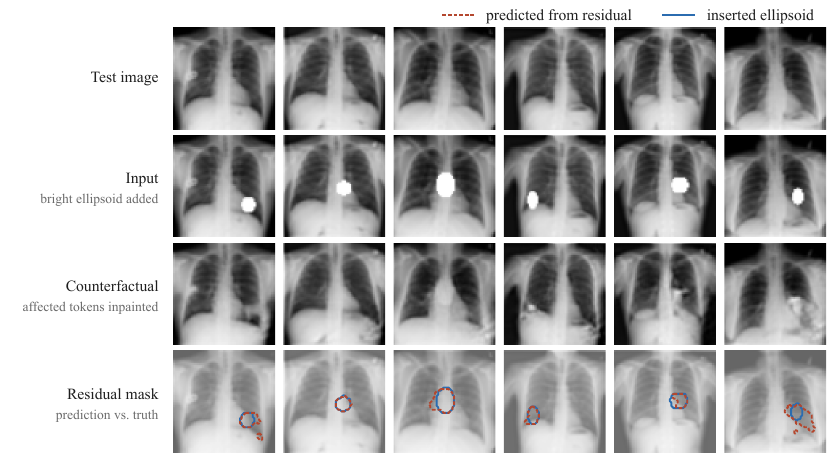}
\caption{Counterfactual inpainting probe with LFQ-1024 + MaskGIT. A bright ellipsoid is added to each test image, the tokens it touches are inpainted, and thresholding the input-counterfactual residual gives the predicted mask, drawn against the inserted ellipsoid in the bottom row. This illustrates a use of the token model and is not evaluated as anomaly detection.}
\label{fig:anomaly}
\end{figure}

For oracle-mask counterfactual inpainting (Fig.~\ref{fig:anomaly}) we paint a synthetic
bright ellipsoid at a random off-centre lung location, encode the
anomalous image, identify which tokens lie within the anomaly region,
and run the same MaskGIT inpainting pipeline on those tokens. This
produces a counterfactual reconstruction in which the model's learned
data prior replaces the anomaly, and a binary anomaly segmentation mask derived by adaptive thresholding of
$|x_{\text{input}} - x_{\text{counterfactual}}|$ with small-component
morphological clean-up. Across the six demo cases, the predicted
segmentation has IoU 0.31--0.96 against the ground-truth ellipsoid mask.
These examples show that the trained model supports oracle-region
inpainting without retraining; they are not a blind
anomaly-localisation or clinical anomaly-detection evaluation. The
same procedure applies to any of the four discrete-diffusion / masked-LM
generators in the library, and the continuous-latent generators (LDM, RF)
admit an analogous SDEdit-style pipeline~\citep{meng2021sdedit}, which is
left to future work.

\paragraph{Demonstrated observations vs.\ proposed mechanisms.}
The demonstrated findings are the quantizer--generator interaction verified
on the three-seed vocabulary-1024 block, the single-seed ChestMNIST-64 grid
for the wider vocabulary sweep, the validation-selected sampler effects, the
weak alignment between PSNR and generation FID-192, and the limited
cross-dataset replication. The neighbour-conditional predictive
gain and marginal token entropy were tested as generator-free correlates
of modelability, and neither tracks generation quality. The remaining explanations (hypercube structure, VQ
position-specific structure, latent-scale sensitivity, and quantization or
decoding contributions to nearest-neighbour distance) remain untested
hypotheses. Direct tests
would require reconstruction FID gaps, larger seed counts, conditional
token-prediction metrics, dependency measures, and uniform continuous-latent
normalisation ablations.

\subsection{Limitations and Broader Impact}

Nine limitations are listed below.
\textbf{(i) Limited dataset coverage, single resolution, primarily one modality.}
Most quantitative results in this paper are on ChestMNIST at
$64 \times 64$; the cross-dataset replication on PneumoniaMNIST and
OrganAMNIST above is a partial validation, but full generalisation claims (vocabulary
advantage, D3PM/SEDD sampler sensitivity, reconstruction--generation decoupling) require
further replication across the MedMNIST family and at higher resolutions.
The work is therefore presented as a controlled case study, and the libraries and protocols are released
so that others can test broader generalisation.
\textbf{(ii) Seeded interaction block, single-seed wider grid.} The
vocabulary-1024 interaction block ($\{$VQ,LFQ,FSQ$\}\times\{$AR,MaskGIT,D3PM$\}$
plus LFQ+SEDD) is trained at three seeds (42/43/44; median per-cell std
0.047, Table~\ref{tab:seed_variance}), and the seed-verified interaction
claim applies only to that block. The wider grid at vocabularies 2{,}048 and
4{,}096 remains single-seed, so its per-cell variance and rank stability are
unestimated, and fine-grained rankings there (e.g.\ LFQ-1024 + AR vs
FSQ-1024 + AR at FIDs 0.33 vs 0.39, which is within seed noise even in the
seeded block) should not be over-read. The c8 continuous anomaly in
particular cannot be attributed to architecture with confidence without a
multi-seed re-run.
\textbf{(iii) FID-192 is not directly comparable to FID-2048
literature.} We use 192-dimensional InceptionV3 features
(Appendix~\ref{app:implementation}); this is a convenience choice for the
low-resolution regime and is not a standard benchmark, and absolute FID
values do not translate to natural-image FID-2048 numbers. We verified that standard
FID-2048 is usable at our sample size and that it rank-agrees with FID-192
(Spearman 0.89; Table~\ref{tab:fid_dual}), and a classifier two-sample test
that uses neither Inception features nor a Gaussian assumption agrees at 0.86,
so the ranking is not an artifact of the feature layer. The specific
mid-range orderings on which the two metrics disagree (AR LFQ-vs-FSQ and the
best mid-range MaskGIT cell) are left unranked under both metrics.
\textbf{(iv) Unconditional generation only.} ChestMNIST has 14 binary
pathology labels that we do not condition on; class-conditional
extensions remain untested.
\textbf{(v) No clinical evaluation.} We rely on FID-192 for ranking (with
Inception Score computed but, as App.~\ref{app:extended} explains, uninformative for
ranking), with no radiologist evaluation of generated samples and no
synth-then-classify utility experiment.
\textbf{(vi) Finite HP grid.} Tuned values in Table~\ref{tab:hp_sweep} are
selected on a held-out validation split and evaluated once on test, so they
are not test-selected; residual optimism is bounded by the finite sweep grid
and the validation screening noise floor (0.010 at 2K), and a larger grid
could shift the selected operating points. The reduced-step D3PM/SEDD
results therefore describe the implemented sampler and sweep grid and do
not constitute a general result about D3PM respacing.
\textbf{(vii) Dataset and label bias.} ChestMNIST inherits the
sampling, acquisition, and label-noise biases of its source chest-X-ray
corpus as packaged by MedMNIST~\citep{yang2023medmnist}. The labels are
not a substitute for radiologist-adjudicated clinical truth, and the
$64 \times 64$ downsampled images remove many subtle findings that
would matter in deployment. Because our generators are unconditional,
they learn the marginal image distribution rather than calibrated
pathology-conditional distributions; generated samples should therefore
not be interpreted as clinically representative, balanced, or suitable
for patient-facing use without additional auditing.
\textbf{(viii) Evaluation-metric dependence.} Our headline rankings rest on
FID-192 in an ImageNet-trained feature space; although we cross-check against
a domain-trained ChestMNIST feature space (App.~\ref{app:fid_validation}), we
do not report a radiomics-based generative metric such as the Fr\'echet
Radiomic Distance~\citep{konz2024frd}, reconstruction FID, or
generator-to-reconstruction FID gaps in the same feature spaces, and feature-extractor
choice is known to affect medical generative-model evaluation~\citep{woodland2024miccai}.
\textbf{(ix) Privacy is probed only by nearest neighbours.} The memorization
probes are feature-space, pixel-space, and LPIPS nearest-neighbour checks;
they do not test membership inference or patient-level duplication, which
would require training-set membership labels or patient identifiers that we
do not use here.

\paragraph{Broader impact.}
Both \medtok{} and \medlat{} are public libraries. The
intended use is methodological: controlled comparison of medical-image
tokenizers and latent generators. The release should not be used to
train or deploy clinical decision systems without dataset-specific
governance, privacy review, and external clinical validation. Specific
misuse risks include presenting synthetic scans as real or as
de-identified real data, using generated images to inflate or launder
datasets, and amplifying the demographic and acquisition biases of the
source corpus; none of these are mitigated by the release itself.

\section{Reproducibility}
\label{sec:availability}

\paragraph{Software.} The code is released as two MIT-licensed libraries, available on GitHub and PyPI, and a paper repository:
\begin{itemize}[leftmargin=*,itemsep=2pt]
    \item \textbf{\medtok{}}: tokenizer training, reconstruction, and tokenization, \url{https://github.com/liamchalcroft/medtokenizers}
    \item \textbf{\medlat{}}: latent-generator training, sampling, and evaluation, \url{https://github.com/liamchalcroft/medlatents}
    \item \textbf{tokenizer-generator-coupling}: the experiment scripts, result files and figure scripts for this paper, \url{https://github.com/liamchalcroft/tokenizer-generator-coupling}
\end{itemize}
The paper repository pins both libraries to released versions in a locked
environment. The bootstrap, domain-FID, throughput, factor-decomposition
and nearest-neighbour memorization utilities used here are part of
\medlat{}, and App.~\ref{app:library} summarises the library interfaces.
The L40S runs (the LFQ and FSQ seed replicates, the tuned multi-seed runs, the LFQ side of the data-scale control and the LFQ and FSQ FID-2048 D3PM rows) used an environment rebuilt from the same lockfile; re-running LFQ-1024 + D3PM in it moves the default FID-192 from 0.44 to 0.43 and the tuned value from 0.0898 to 0.0885.

\paragraph{Data.} ChestMNIST, PneumoniaMNIST, and OrganAMNIST are taken from
the public MedMNIST v2 package~\citep{yang2023medmnist} at $64 \times 64$; no
new dataset is released.

\paragraph{Pipeline and configurations.} The factorial sweep and the
sampling-hyperparameter sweep are run from the tokenizer-generator-coupling
repository and read the token files written by \medtok{}. All
tokenizers and generators are trained with seed 42 under the optimiser,
schedule, and capacity settings of \S\ref{sec:tokenizers} and
\S\ref{sec:generators}, and each generator checkpoint records the
latent-normalisation statistics needed to replay sampling.

\paragraph{Compute.} All headline training and evaluation runs used a single
NVIDIA A6000 (48\,GB). Recorded generation/evaluation jobs account for
80.6 GPU-hours across ChestMNIST, PneumoniaMNIST, and OrganAMNIST, including
70.6 GPU-hours for the ChestMNIST default-configuration evaluations and their repeats. Recorded
sampling-hyperparameter sweeps account for a further 53.4 GPU-hours. Training wall clocks
estimated from checkpoint timestamps sum to 156.6 GPU-hours for tokenizers and
265.5 GPU-hours for generators, so the total budget is approximately
556 A6000-hours. These training estimates are upper bounds because they use
wall-clock gaps between checkpoint writes and include per-run overhead. The LFQ and FSQ seed-variance runs and the LFQ side of the data-scale control used a single NVIDIA L40S, and the VQ seed and data-scale runs used the main-grid A6000; neither is included in this budget.

\begin{ack}
This work received no specific funding. The author declares no competing interests.
\end{ack}

\bibliographystyle{plainnat}
\bibliography{references}

\appendix

\section{Extended Results}
\label{app:extended}

\begin{figure}[h]
\centering
\includegraphics[width=\linewidth]{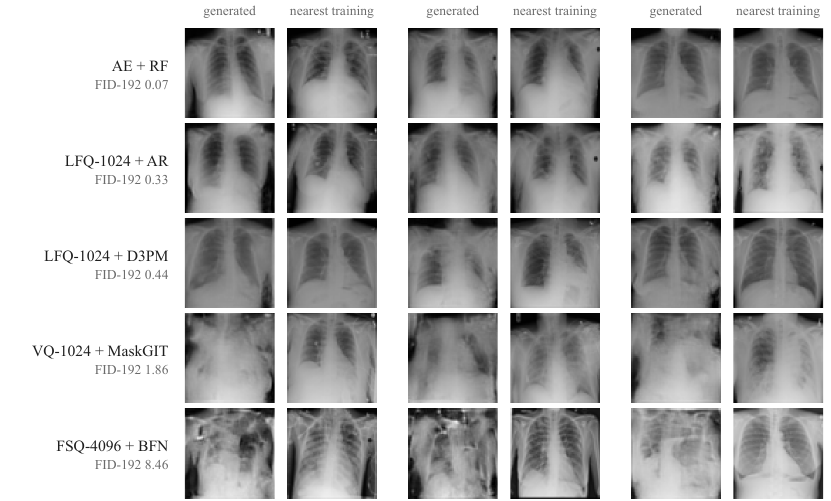}
\caption{The three closest generated-to-training pairs for five cells under the feature-space probe, with the generated sample on the left of each pair.}
\label{fig:memorization_gallery}
\end{figure}

Inception scores at default sampling configurations were computed during evaluation but are omitted from the main tables because their across-cell standard deviation is small (0.16) and they contribute no additional ranking information. Figure~\ref{fig:memorization_gallery} shows the nearest-neighbour gallery for five representative cells from the feature-space memorization probe.

\clearpage

\section{Tokenizer Analysis}
\label{app:tokenizer}

\begin{figure}[h]
\centering
\includegraphics[width=0.82\linewidth]{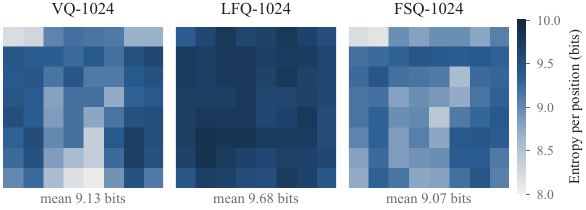}
\caption{Shannon entropy of the token at each of the $8\times8$ latent positions over the ChestMNIST test set, at vocabulary 1024 (maximum 10 bits). Positions are laid out as in the image, so the top row corresponds to the top of the radiograph.}
\label{fig:per_pos_entropy}
\end{figure}

\paragraph{Per-position token entropy.} Figure~\ref{fig:per_pos_entropy} shows the per-position Shannon entropy heatmap for each discrete tokenizer at vocabulary 1{,}024. LFQ achieves uniform near-maximum entropy across all 64 spatial positions; VQ shows mild edge-vs-centre variation; FSQ shows the largest spatial variation, with corner positions noticeably lower-entropy than centre positions. Code-usage histograms (sorted by frequency) are uniform for LFQ and VQ across all three vocabulary sizes (100\% utilisation, near-flat distribution); FSQ's histogram has a heavier tail at vocabulary 4{,}096, consistent with FSQ's lower entropy at larger vocabularies (Table~\ref{tab:tokenizer_props}).

\paragraph{Continuous tokenizer latent statistics.} Across the channel and KL sweeps, the test-split latents have means consistently $|\mu|\!<\!0.1$ but the standard deviations depart from unit variance in both directions: VAE-c16 collapses to $\mathrm{std}\!=\!0.62$ ($|z|_\text{mean}\!=\!0.35$); VAE at strong KL ($\lambda_\text{KL}\!=\!10^{-5}$) reduces to $\mathrm{std}\!=\!0.54$; AE (no KL constraint) inflates to $\mathrm{std}\!=\!2.03$ ($|z|_\text{mean}\!=\!1.60$). The cells with std closest to 1.0 (VAE-c2/c4/c8 and VAE at $\lambda_\text{KL}\!\in\!\{10^{-6},10^{-7}\}$) have the latent scale that LDM's noise schedule assumes (c8's LDM outlier aside); the low-std cells (c16, $10^{-5}$) perform poorly with LDM but not with RF, consistent with the mechanism in \S\ref{sec:generators_results}.

\clearpage

\section{Implementation Details}
\label{app:implementation}

\paragraph{Tokenizer training details.}
All tokenizers are trained with \medtok{} on
ChestMNIST-$64$ with batch size 128, AdamW ($\mathrm{lr}=10^{-4}$,
weight decay 0.01), cosine learning-rate decay over 50 epochs, BFloat16
mixed precision, gradient clipping at 1.0, validation every five epochs,
model EMA decay 0.9999, and best-checkpoint selection by validation total
loss. Inputs are min--max normalised to $[0,1]$ and loaded without data
augmentation. The shared encoder--decoder has $z_\mathrm{channels}=64$,
embedding dimension 16 for discrete tokenizers, base width 64,
channel multipliers $(1,2,4)$, two residual blocks per resolution,
attention at $16\times16$, dropout 0, and spatial compression 8.
All tokenizers share the reconstruction, perceptual, adversarial, and LeCam
terms of \S\ref{sec:tokenizers}; discrete tokenizers add a quantization-loss
term with weight 1.0.
VAE tokenizers use the same reconstruction weight with KL weight
$\lambda_\mathrm{KL}$, linearly warmed up over 10 epochs; AE is the
$\lambda_\mathrm{KL}=0$ endpoint. Table~\ref{tab:tokenizer_hparams} lists
the family-specific quantizer settings.

\begin{table}[h]
\centering
\caption{Tokenizer-specific settings omitted from the main text. VQ uses
EMA codebook updates with cosine-normalised embeddings. FSQ level vectors
define the implicit Cartesian-product vocabulary. LFQ uses one binary
codebook with dimension $\log_2 K$.}
\label{tab:tokenizer_hparams}
\scriptsize
\begin{tabular}{@{}p{0.10\linewidth}p{0.20\linewidth}p{0.62\linewidth}@{}}
\toprule
\textbf{Family} & \textbf{Vocabulary} & \textbf{Quantizer settings} \\
\midrule
VQ & 1{,}024 / 2{,}048 / 4{,}096 &
$K$ embeddings; EMA decay 0.99; $\beta=0.25$; L2-normalised codes; commitment/codebook terms via VQ loss \\
LFQ & 1{,}024 / 2{,}048 / 4{,}096 &
codebook dim 10 / 11 / 12; one codebook; quantization temperature 0.01; entropy weight 0.1; commitment weight 0.25 \\
FSQ & 1{,}024 &
levels $(4,4,4,4,4)$; no auxiliary quantizer loss \\
FSQ & 2{,}048 &
levels $(8,4,4,4,4)$; no auxiliary quantizer loss \\
FSQ & 4{,}096 &
levels $(8,8,4,4,4)$; no auxiliary quantizer loss \\
\bottomrule
\end{tabular}
\end{table}

\paragraph{Continuous latent extraction.}
After tokenizer training, each image is encoded once with the tokenizer
in evaluation mode and stored as a float16 array. For the VAE configuration
used here, the encoder returns the posterior mean at evaluation time; generator training therefore uses a fixed posterior-mean
latent per image rather than resampling a posterior draw each epoch. AE
latents are deterministic. Continuous generators then read those stored
latents from disk; latent normalisation, when enabled, is computed only
from the training split and saved with the generator checkpoint.

\paragraph{Reduced-step D3PM/SEDD sampling.}
The sampling sweep in Table~\ref{tab:hp_sweep} does not retrain the
D3PM or SEDD checkpoints and does not use an arbitrary-jump posterior
over the original 1{,}000-step chain. For each candidate
$N\in\{100,250,500,1000\}$, the sampler constructs a fresh $N$-step
absorbing process with the same cosine schedule family:
$\{\beta_n,\bar{\alpha}_n,\bar{\alpha}_{n-1}\}_{n=0}^{N-1}$.
Sampling starts from the all-mask state and applies adjacent reverse
steps for this $N$-step process:
\[
x \leftarrow \mathrm{MASK}^{64}, \qquad
\text{for } n=N-1,\ldots,0:\quad
x \leftarrow p_\theta(x_{n-1}\mid x_n=x,n;\beta_{0:N-1}).
\]
For masked positions, the matrix-free absorbing step uses the model
prediction $p_\theta(x_0=j\mid x_n,n)$ and samples
\[
p(x_{n-1}=j\mid x_n=\mathrm{MASK})
= p_\theta(x_0=j\mid x_n,n)
  \frac{\beta_n\bar{\alpha}_{n-1}}{1-\bar{\alpha}_n}, \quad j<K,
\]
\[
p(x_{n-1}=\mathrm{MASK}\mid x_n=\mathrm{MASK})
=\frac{1-\bar{\alpha}_{n-1}}{1-\bar{\alpha}_n},
\]
followed by normalisation; unmasked positions persist deterministically.
For SEDD, the same sampler is used but the model receives a
normalised time $n/N$, matching the continuous-time training interface.

\paragraph{Sampling-hyperparameter grids and selected settings.}
The validation sweep used the following candidate grids: AR
temperature $\{0.5,0.7,0.8,0.9,1.0,1.2\}$ and
$k\in\{0,64,256\}$; MaskGIT temperature
$\{0.5,0.7,0.9,1.0,1.2\}$, steps $\{4,8,12,16,24,32\}$, and
mask schedule \{cosine, linear\}; DFM temperature
$\{0.5,0.7,0.9,1.0,1.2\}$ and Euler steps
$\{10,25,50,100,200\}$; D3PM/SEDD temperature
$\{0.5,0.7,0.9,1.0,1.2\}$ and timesteps
$\{100,250,500,1000\}$; BFN temperature
$\{0.3,0.5,0.7,0.9,1.0,1.2\}$ and steps
$\{10,25,50,100,200\}$. Table~\ref{tab:selected_hps}
lists the validation-selected settings that produce
Table~\ref{tab:hp_sweep}.

\begin{table}[h]
\centering
\caption{Validation-selected sampling hyperparameters at vocabulary 1{,}024.
$T$ denotes sampling temperature; $k=0$ means no top-$k$ truncation; $N$
denotes D3PM/SEDD timesteps. Hyperparameters were selected on the 2K
validation screen and evaluated once on 10K test samples.}
\label{tab:selected_hps}
\scriptsize
\begin{tabular}{@{}lllr@{}}
\toprule
\textbf{Tokenizer} & \textbf{Generator} & \textbf{Selected HP} & \textbf{Test FID} \\
\midrule
LFQ-1024 & AR & $T=0.9,\ k=0$ & 0.33 \\
LFQ-1024 & MaskGIT & $T=1.2,\ \mathrm{steps}=4,\ \mathrm{linear}$ & 1.89 \\
LFQ-1024 & DFM & $T=0.7,\ \mathrm{Euler}=200$ & 1.58 \\
LFQ-1024 & D3PM & $T=0.7,\ N=100$ & 0.09 \\
LFQ-1024 & SEDD & $T=0.7,\ N=500$ & 0.10 \\
LFQ-1024 & BFN & $T=1.2,\ \mathrm{steps}=100$ & 2.13 \\
FSQ-1024 & AR & $T=1.0,\ k=0$ & 0.29 \\
FSQ-1024 & MaskGIT & $T=0.9,\ \mathrm{steps}=32,\ \mathrm{linear}$ & 0.88 \\
FSQ-1024 & DFM & $T=0.5,\ \mathrm{Euler}=200$ & 1.47 \\
FSQ-1024 & D3PM & $T=1.0,\ N=250$ & 0.13 \\
FSQ-1024 & SEDD & $T=1.0,\ N=250$ & 0.10 \\
FSQ-1024 & BFN & $T=1.2,\ \mathrm{steps}=100$ & 5.80 \\
VQ-1024 & AR & $T=1.2,\ k=0$ & 0.46 \\
VQ-1024 & MaskGIT & $T=1.2,\ \mathrm{steps}=8,\ \mathrm{cosine}$ & 1.42 \\
VQ-1024 & DFM & $T=0.5,\ \mathrm{Euler}=100$ & 3.13 \\
VQ-1024 & D3PM & $T=0.5,\ N=500$ & 0.15 \\
VQ-1024 & SEDD & $T=0.7,\ N=250$ & 0.13 \\
VQ-1024 & BFN & $T=1.2,\ \mathrm{steps}=100$ & 4.10 \\
\bottomrule
\end{tabular}
\end{table}

\paragraph{Matrix-free absorbing D3PM.}
Standard D3PM~\citep{austin2021d3pm} precomputes three transition matrices ($\mathbf{Q}_t$, $\bar{\mathbf{Q}}_t$, and the shifted $\bar{\mathbf{Q}}_{t-1}$), each in $\mathbb{R}^{(K+1) \times (K+1)}$ for all $T$ timesteps, requiring $O(TK^2)$ transition-matrix storage: about 201\,GB at $K\!=\!4{,}096$ and $T\!=\!1{,}000$ in float32, far beyond a 48\,GB GPU. For absorbing transitions, the posterior $q(x_{t-1} \mid x_t, x_0)$ has closed form: unmasked tokens deterministically persist ($x_{t-1} = x_t$), while masked tokens are drawn from a two-point distribution over $\{x_0, \text{MASK}\}$ with probabilities proportional to $\beta_t \bar{\alpha}_{t-1}$ and $1 - \bar{\alpha}_{t-1}$ respectively. The implementation used here reduces transition-matrix storage to $O(T)$ and computation per step to $O(BLK)$ where $B$ is batch size, $L$ is sequence length, and $K$ is vocabulary size; total training memory still includes vocabulary-dependent logits, embeddings, activations, and optimiser state. This enables D3PM training on vocabularies up to 64{,}000 on a single 48\,GB GPU.

\paragraph{FID-192 for small medical images.}
Standard FID uses InceptionV3's 2{,}048-dimensional final average-pooling features. We use the \texttt{torchmetrics} implementation~\citep{torchmetrics} with 192-dimensional features and inputs scaled to $[0,1]$; these features globally average the second max-pooling InceptionV3 block. FID-192 is a convenience choice for this low-resolution regime. Standard FID-2048 is also usable here, with a real-vs-real floor at 10K samples of 2.31 (std 0.016, no negative values across 20 resamples), and the two metrics rank-agree at Spearman $\rho = 0.89$ (Table~\ref{tab:fid_dual}). FID-192 is adopted as the internal ranking metric because its lower real-vs-real floor ($\sim$0.002 at 10K) gives better separation among the strong cells. Its absolute values are not comparable to standard FID-2048, and neither metric establishes clinical validity.

\section{FID-192 Domain-Feature Sanity Check}
\label{app:fid_validation}

To check that the low-resolution InceptionV3 feature choice does not
drive the conclusions, we recomputed FID on the six cross-dataset
replication cells in a domain-trained feature space: the 512-dimensional
penultimate layer of a ResNet-18 multi-label ChestMNIST classifier
trained on the real training split. The classifier reaches mean test
AUC 0.761 across 14 pathologies, so its representation is domain-relevant
for ChestMNIST but independent of the generator/tokenizer training objective.
Table~\ref{tab:domain_fid_validation} compares FID-192
against this domain-FID. The rank correlation is $\rho=0.71$ (exact
two-sided permutation $p=0.14$ over the six rankings), which is consistent
with FID-192 but too small a check to confirm it on its own. Both metrics
put the continuous AE + RF and VAE-c2 + LDM cells first; the disagreements
are among the three AR cells, whose domain-FIDs lie within 0.6 of each
other, and between VQ-1024 + AR and LFQ-1024 + D3PM.

\begin{table}[h]
\centering
\caption{Domain-FID sanity check on the six cross-dataset replication
cells. Domain-FID uses 512-dimensional features from a real-data
ChestMNIST ResNet-18 classifier (mean test AUC 0.761). Lower is better;
$n_\text{gen}=10{,}000$, $n_\text{real}=22{,}433$. Spearman
$\rho=0.71$ with exact permutation $p=0.14$.}
\label{tab:domain_fid_validation}
\small
\begin{tabular}{@{}lcc@{}}
\toprule
\textbf{Cell} & \textbf{FID-192} & \textbf{Domain-FID} \\
\midrule
LFQ-1024 + AR        & 0.33 & 17.08 \\
FSQ-1024 + AR        & 0.39 & 16.53 \\
VQ-1024 + AR         & 0.52 & 16.93 \\
LFQ-1024 + D3PM      & 0.44 & 24.70 \\
VAE-c2 + LDM         & 0.14 & 7.67 \\
AE + RF          & 0.07 & 0.74 \\
\bottomrule
\end{tabular}
\end{table}

\paragraph{Standard FID-2048 anchor.} To check that FID-192 is not a
feature-layer artifact, we recomputed standard FID-2048 from the same
generated samples on 12 cells spanning the full quality range (0.07 to 8.44).
FID-2048 is not
degenerate at this resolution (App.~\ref{app:implementation}). FID-192 and FID-2048 rank-agree at Spearman 0.89
across the 12 cells and agree on the coarse structure and the best cells
(AE + RF, tuned LFQ-1024 + D3PM, and the VAE references are lowest under both), and
they disagree on the order of some weaker cells and on the LFQ-versus-FSQ pair
under AR. Table~\ref{tab:fid_dual} shows the vocabulary-1024
interaction block under both metrics. Under FID-2048 the best quantizer still
changes with the generator (FSQ under AR, LFQ under MaskGIT and D3PM), so the
non-separability holds under both metrics and no quantizer can be ranked
independently of the generator. The D3PM interaction is metric-robust, with
the ordering LFQ $<$ FSQ $<$ VQ identical under both metrics. Two mid-range orderings are metric-dependent and are left
unranked, namely the swap between LFQ and FSQ under AR (already reported as
within seed noise) and the disagreement between the two metrics on which
poor MaskGIT cell is best.

\begin{table}[h]
\centering
\caption{Vocabulary-1024 interaction block under FID-192 and standard FID-2048
(10K samples, identical generated images for both metrics). The best quantizer
per generator changes under both metrics (non-separability). FID-192 values are
recomputed from this generation pass rather than carried over from
Table~\ref{tab:main_results}, so they differ by up to 0.07 (e.g.\ FSQ+D3PM 1.14 here vs.\ 1.21 there, VQ+MaskGIT 1.83
here vs.\ 1.86 there). The VQ+D3PM FID-2048 value was computed on the CPU, which
reproduces the GPU FID-2048 of a reference cell to within 0.01.}
\label{tab:fid_dual}
\small
\begin{tabular}{@{}l cc cc cc@{}}
\toprule
& \multicolumn{2}{c}{\textbf{VQ-1024}} & \multicolumn{2}{c}{\textbf{LFQ-1024}} & \multicolumn{2}{c}{\textbf{FSQ-1024}} \\
\cmidrule(lr){2-3}\cmidrule(lr){4-5}\cmidrule(lr){6-7}
\textbf{Generator} & FID-192 & FID-2048 & FID-192 & FID-2048 & FID-192 & FID-2048 \\
\midrule
AR      & 0.53 & 59  & 0.34 & 52  & 0.41 & 35 \\
MaskGIT & 1.83 & 217 & 1.94 & 79  & 1.16 & 194 \\
D3PM    & 1.30 & 114 & 0.42 & 42  & 1.14 & 59 \\
\bottomrule
\end{tabular}
\end{table}

\paragraph{A label-free two-sample check.} Because our generators are
unconditional, a synth-then-classify utility study is ill-posed (no class
labels to train a downstream classifier), so as a label-free proxy we ran a
classifier two-sample test: a real-vs-generated discriminator AUC (three
seeds) over 12 cells, which uses neither Inception features nor a Gaussian
assumption. Its ranking agrees with FID-192 at Spearman 0.86, which provides
a second independent check alongside the weaker domain-feature FID agreement
(0.71) above. The best cell (AE + RF) reaches AUC 0.54, so a discriminator can barely
separate its samples from real images, while the poor cells sit near 1.0, in
line with their FID. A proper downstream-utility assessment requires
pathology-conditional generation followed by synth-then-classify.

\section{FID Bootstrap Confidence Intervals}
\label{app:fid_ci}

We report estimator uncertainty for four representative cells using a
paired bootstrap over real and generated feature indices. For each cell
we generated 10{,}000 samples, extracted FID-192 features, resampled
10{,}000 real indices and 10{,}000 generated indices with replacement,
and recomputed FID for $B=200$ bootstrap replicates. These intervals
measure FID estimator noise for a fixed trained generator; they do not
estimate train-seed variance. Table~\ref{tab:fid_ci} shows that the
intervals are narrow relative to the coarse quality regimes in
Table~\ref{tab:main_results}.

\begin{table}[h]
\centering
\caption{Paired-bootstrap 95\,\% confidence intervals for representative
FID-192 cells ($B=200$, 10K samples per bootstrap replicate).}
\label{tab:fid_ci}
\small
\begin{tabular}{@{}lcccc@{}}
\toprule
\textbf{Cell} & \textbf{Mean} & \textbf{Std.} & \textbf{2.5\%} & \textbf{97.5\%} \\
\midrule
AE + LDM          & 0.093 & 0.004 & 0.086 & 0.101 \\
LFQ-1024 + AR         & 0.337 & 0.009 & 0.320 & 0.354 \\
LFQ-1024 + MaskGIT    & 1.938 & 0.029 & 1.876 & 1.988 \\
FSQ-4096 + BFN        & 8.441 & 0.077 & 8.281 & 8.597 \\
\bottomrule
\end{tabular}
\end{table}

\section{Library Architecture}
\label{app:library}

Both libraries follow a configure--train--sample--evaluate workflow.
\medtok{} builds discrete and continuous tokenizers on a shared
encoder--decoder, with the quantizer and its size set in the configuration.
\medlat{} implements the eight generator families evaluated here on shared
discrete and continuous DiT backbones and an autoregressive transformer.
The two packages are
connected by a shared token-file format, so any \medtok{} tokenizer can be
paired with any \medlat{} generator. The full pipeline (pre-train tokenizer
$\to$ tokenize dataset $\to$ train generator on tokens $\to$ generate $\to$
FID) covers the 54 discrete cells and 16
continuous-latent reference cells evaluated here.

\end{document}